\documentclass[letterpaper]{article} 
\usepackage{aaai2027}

\usepackage[hyphens]{url}  
\usepackage{graphicx} 
\usepackage{natbib}  
\usepackage{caption} 
\usepackage{amsmath}
\usepackage{algorithm}
\usepackage{algorithmic}
\usepackage{amsfonts}
\usepackage{booktabs}
\usepackage{enumitem}
\usepackage{MnSymbol} 
\usepackage{bm}
\usepackage{dsfont}
\usepackage{multirow}
\usepackage{subfigure}
\usepackage{diagbox}
\usepackage{pifont}
\usepackage{xcolor}
\usepackage{xspace}
\usepackage{colortbl}
\usepackage{listings}

\DeclareCaptionStyle{ruled}{labelfont=normalfont,labelsep=colon,strut=off} 
\floatstyle{ruled}
\newfloat{listing}{tb}{lst}{}
\floatname{listing}{Listing}

\newcommand{\m}{SymboUQ\xspace}
\newcommand{\graycell}[1]{\textcolor{black!60}{#1}}

\title{SymboUQ: Symbolic Uncertainty Quantification for Spatial Reasoning in LLMs}

\author{
Dahai Yu,
Lin Jiang,
Rongchao Xu,
Guang Wang\thanks{Prof. Guang Wang is the corresponding author.}
}
\affiliations{
Florida State University, Tallahassee, Florida\\
dahai.yu@fsu.edu, lin.jiang@fsu.edu, rxu@fsu.edu, guang@cs.fsu.edu \\
}

\begin{document}

\maketitle

\begin{abstract}
Although large language models (LLMs) can produce fluent spatial reasoning traces, their intermediate relations may fail to support the final conclusion, making token-level confidence insufficient for final-answer reliability estimation. Existing formal verifiers provide stronger semantic evidence, but their applicability is partial: a parsed claim need not yield a definite semantic verdict. To address this issue, we introduce \m, a symbolic uncertainty quantification framework that estimates final-answer reliability from reasoning traces by distinguishing \emph{symbolizability}, whether a claim can be represented in the verifier's formal language, from \emph{semantic determinacy}, whether its execution yields an entailed or contradicted verdict rather than an unknown or not-evaluable outcome. \m comprises (i) a Layout Auditor that executes ordered spatial claims and extracts feasibility, conflict, and repair evidence; (ii) a label-free Determinacy Profile that characterizes effective executable coverage; and (iii) a Determinacy-Aware Reliability Composer that integrates constraint-based, representation-based, and decoding-based scores according to verifier applicability. 
Extensive experiments on five spatial reasoning benchmarks with four frozen LLM backbones show that \m achieves approximately an $8\%$ relative improvement in AUROC and a $7\%$ relative reduction in class-balanced Brier loss over the strongest baseline.
\end{abstract}

\section{Introduction}\label{sec:introduction}

Large language models (LLMs) are increasingly used to reason over textual descriptions of environments, yet a plausible final answer may conceal an invalid chain of relational reasoning. This problem is especially acute in spatial reasoning, where a single incorrect directional relation, containment relation, or state transition can corrupt the inferred spatial configuration while leaving the generated explanation fluent and seemingly coherent. We therefore estimate final-answer reliability using both token-level likelihoods and evidence that the generated reasoning supports its conclusion.

Existing uncertainty quantification (UQ) methods estimate reliability from signals such as verbalized confidence, token-level statistics, repeated generations, semantic agreement, and internal representations \citep{kadavath-etal-2022-know-what,manakul-etal-2023-selfcheckgpt,kuhn-etal-2023-semantic-uncertainty,azaria-mitchell-2023-internal-state}. Moving beyond output-level signals, process supervision and learned step-level verifiers assess intermediate reasoning steps and provide more fine-grained estimates of correctness \citep{cobbe-etal-2021-training-verifiers,lightman-etal-2023-verify-step}. Formal verifiers provide complementary evidence by explicitly applying predefined rules to generated claims \citep{suresh-etal-2025-beaver,he-etal-2026-doverifier,manchingal-etal-2026-energy}. However, statistical and learned signals may correlate with correctness without directly evaluating the underlying spatial semantics, whereas formal verification applies only when claims can be reliably parsed, grounded, and executed. Even after a spatial relation is successfully extracted, a verifier may be unable to determine whether it is entailed or contradicted because the referenced entities are ungrounded, the available premises are insufficient, or an earlier generated claim has rendered the reasoning context inconsistent. Consequently, parse coverage alone does not indicate how much usable evidence a formal verifier provides.

To characterize verifier applicability more precisely, we distinguish between two properties that parse coverage fails to separate: \emph{symbolizability} and \emph{semantic determinacy}. Symbolizability refers to whether a generated claim can be translated into the verifier's formal language, whereas semantic determinacy refers to whether executing the translated claim yields a definite verdict of entailment or contradiction rather than an unknown or not-evaluable outcome. For example, a relation may be translated into a valid spatial constraint but remain semantically indeterminate when the trusted scene provides insufficient information to support or refute it. This distinction identifies two complementary evidence regimes. Constraint-based evidence is most informative when a substantial portion of a reasoning trace is semantically determinate. When many claims cannot be conclusively resolved, even if they are symbolizable, neural and decoding-based signals may provide valuable complementary evidence.

Leveraging these complementary evidence regimes to estimate final-answer reliability presents \textbf{three challenges}. First, reliable estimation depends on executing generated claims sequentially while keeping the standing of each claim explicit, so that an unparsed or semantically unresolved claim contributes no positive evidence and any verdict that rests on earlier generated hypotheses is reported as conditional rather than grounded. Second, meaningful use of the execution results calls for a label-free profile that captures symbolizability and semantic determinacy separately from verdict polarity, while explicitly accounting for unresolved claims and claims rendered not evaluable by earlier inconsistencies. Third, the integration of heterogeneous reliability signals should account for variation in the informativeness of symbolic and neural evidence across claims and reasoning traces.

To address these challenges, we propose \m, a symbolic UQ framework for final-answer reliability from spatial reasoning traces. \m comprises three key components. First, the \emph{Layout Auditor} compiles the trusted scene description and ordered reasoning trace into a spatial constraint program. It separates the trusted scene from the hypotheses the trace introduces, assigns each generated claim one of four outcomes relative to an explicitly stated context: entailed, contradicted, unknown, or not evaluable, and extracts evidence related to feasibility, conflicts, and possible repairs. Second, the \emph{Determinacy Profile} summarizes the effective executable coverage of the trace, including the degree of semantic resolution and the loss of coverage caused by an infeasible context. Third, the \emph{Determinacy-Aware Reliability Composer (DARC)} uses a small labeled target-domain validation split to select useful frozen base scores and align their directions. It integrates constraint-, representation-, and decoding-based evidence through applicability-conditioned interactions to estimate final-answer correctness. 

Extensive experiments across five spatial reasoning benchmarks and four frozen LLM backbones demonstrate that \m achieves the strongest overall performance in trace ranking and class-balanced Brier loss under a shared validation-fit protocol. Controlled ablations show that applicability-aware interactions provide consistent improvements beyond score-only target adaptation. Trace-level analyses further show that the utility of constraint-based scores is more closely associated with semantic determinacy than with parse coverage.

The key contributions of this work are as follows:
\begin{itemize}[leftmargin=*,nosep]
    \item Conceptually, we distinguish textual symbolizability from semantic determinacy, providing a principled basis for determining when verifier-derived evidence from a reasoning trace is informative for estimating final-answer reliability.
    
    \item Technically, we propose \m, a symbolic UQ framework comprising three key components: a spatial Layout Auditor, a label-free execution-derived Determinacy Profile, and a Determinacy-Aware Reliability Composer that integrates heterogeneous reliability signals according to effective verifier applicability.
    
    \item Empirically, across five benchmarks and four LLM families, \m averages approximately $8\%$ higher AUROC and $7\%$ lower class-balanced Brier loss than the strongest external baseline. 
\end{itemize}



\section{Problem Formulation}\label{sec:preliminary}

Let $x=(\eta,q)$ denote an input consisting of a scene description $\eta$ and a question $q$. Given $x$, a frozen language model $M$ generates a reasoning trace $r=M(x)$ containing $L$ tokens. We decompose the trace into an ordered sequence of claims
\begin{equation}
\mathcal C(r)=\big((c_k,t_k,m_k)\big)_{k=1}^{K},
\end{equation}
where $c_k$ is the $k$-th claim, $t_k\in\{\textsc{reasoning},\textsc{conclusion}\}$ indicates its type, and $m_k\in\{0,1\}^{L}$ is a binary mask identifying its token span.  The final claim is designated as the conclusion.
Claim order is essential because the validity and evaluability of $c_k$ depend on both the scene description and the preceding claims $(c_1,\ldots,c_{k-1})$.

Let $y\in\{0,1\}$ indicate whether the final conclusion matches the benchmark answer after deterministic normalization. Our goal is final-answer reliability estimation,
\begin{equation}
p(r)\approx \Pr(y=1\mid x,r).
\end{equation}
We consider a target-adapted final-answer reliability setting in which a labeled validation split from the target domain may be used to fit the reliability layer. The generator and any source-trained base estimators remain fixed during target adaptation, and test labels are unavailable for model selection, fitting, and calibration.

\section{Methodology}\label{sec:method}

\begin{figure*}[t]
\centering
\includegraphics[width=\textwidth]{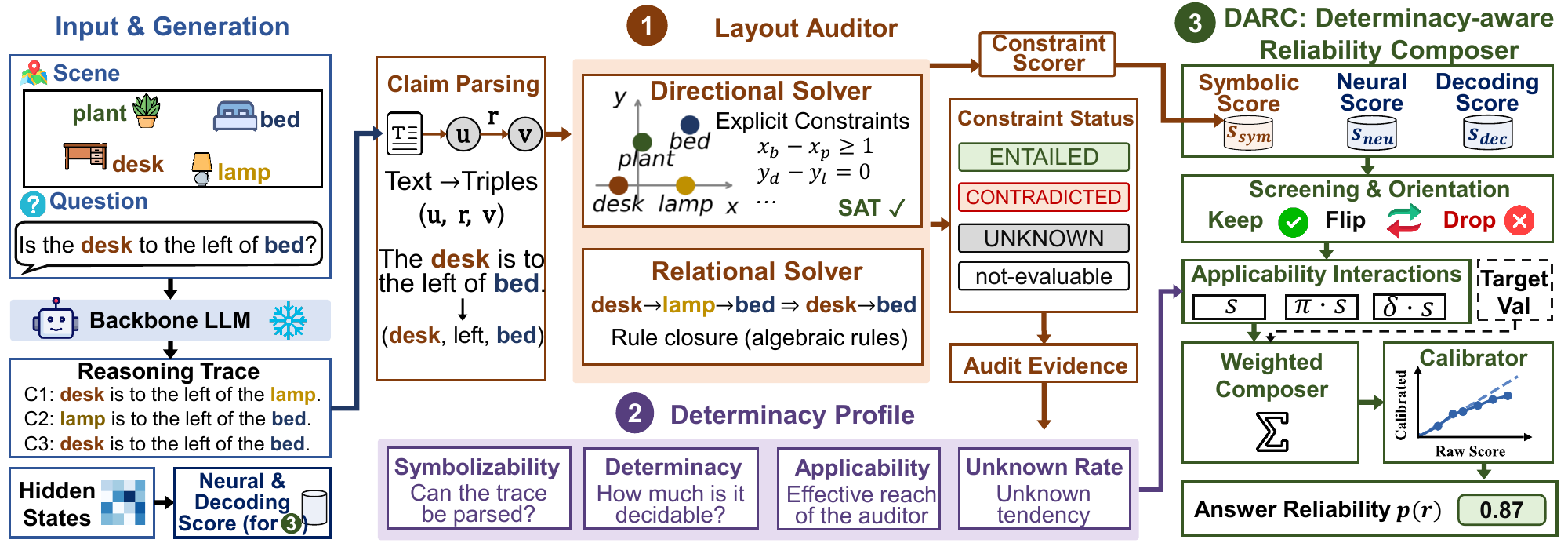}
\caption{Overall framework of \m. The Layout Auditor executes ordered spatial claims and produces constraint-based evidence; the Determinacy Profile measures symbolizability and effective semantic resolution; DARC then conditions heterogeneous final-answer reliability signals on this trace-level execution profile.}
\label{fig:framework}
\end{figure*}

Figure~\ref{fig:framework} shows the overall framework of \m, which consists of three key modules addressing the three challenges described in the Introduction. The Layout Auditor sequentially executes generated claims while ensuring that missing or unresolved coverage is not treated as evidence of correctness. The Determinacy Profile summarizes the resulting execution state without incorporating verdict polarity, and the Determinacy-Aware Reliability Composer (DARC) integrates constraint-based evidence with frozen representation-based and decoding-based estimators. The profile therefore serves as a label-free characterization of effective executable coverage rather than as a standalone confidence score. \textbf{Appendix \ref{app:illustrative_example} provides an illustrative example.}

\subsection{Layout Auditor: From Claims to Constraints}

The Layout Auditor instantiates partial symbolic verification by mapping supported natural-language statements into a canonical relation system. Each recognized statement is represented as a triple $z=(u,\rho,v)$, where $u$ is the subject, $\rho$ is the relation, and $v$ is the object. The ontology covers direction, containment, exclusion, distance, topology, location, and path endpoints. We distinguish the directional \textit{same-position} relation from topological \textit{overlap}. Surface variants, including cardinal-direction terms, clock-face expressions, and containment paraphrases, are normalized to canonical relations. The parser also records its confidence and whether both entities are grounded in the scene.

Let $D$ denote the multiset of relations parsed from the trusted scene description, and let $C_k$ denote the multiset of relations parsed from claim $c_k$. If a claim contains no supported relation, then $C_k=\varnothing$. Such a claim remains unresolved and is never treated as consistent. For state-update tasks, the auditor retains only the most recent location associated with each subject. 

\paragraph{Directional solver.}
Each directional relation is represented by $(\Delta_x,\Delta_y)\in\{-1,0,1\}^2$. For coordinate $a\in\{x,y\}$, a positive component imposes $a_u-a_v\geq1$, a negative component imposes $a_u-a_v\leq-1$, and a zero component imposes equality. Bellman-Ford negative-cycle detection determines if the resulting difference constraints admit a joint layout.

\paragraph{Relational solver.}
The relational solver applies explicit inverse, symmetry, transitivity, incompatibility, containment, and functional-location rules. Appendix~\ref{app_auditor} gives the complete relation families and closure conventions.

Together, the directional and relational solvers define a feasibility predicate $\operatorname{SAT}(B)$ for any relation set $B$. The predicate holds only when both the coordinate-constraint system and the relational closure are feasible.

\paragraph{Three-valued constraint evidence.}
We distinguish three semantic verdicts from a separate evaluability flag:
\begin{equation}
\begin{aligned}
\mathcal T &= \{\textsc{en},\textsc{co},\textsc{un}\},\\
\textsc{en} &= \textsc{entailed}, & \textsc{co} &= \textsc{contradicted},\\
\textsc{un} &= \textsc{unknown}. 
\end{aligned}
\label{eq:status_space}
\end{equation}
In addition, $\textsc{no} = \textsc{not-evaluable}\notin\mathcal T$ means that an infeasible preceding context prevents evaluation; it is not a fourth semantic verdict.

\paragraph{Trusted and assumed contexts.}
A reasoning trace supplies two kinds of premises. The \emph{grounded context} contains the trusted scene together with earlier claims resolved as entailed under grounded replay,
\begin{equation}
T_k=D\cup\bigcup_{\substack{j<k:\\ \tau^T_j=\textsc{en}}}C_j .
\label{eq:trusted}
\end{equation}
The \emph{assumed context} additionally admits every earlier symbolizable claim as a working hypothesis,
\begin{equation}
A_k=D\cup\bigcup_{j<k}C_j,
\qquad T_k\subseteq A_k .
\label{eq:prefix}
\end{equation}
A verdict under $T_k$ follows from grounded information, whereas one under $A_k$ measures coherence with the trace's intermediate commitments. We instantiate the active solver context as $B_k=A_k$, so reported verdicts are conditional on preceding generated hypotheses. This choice exposes self-contradiction but does not turn assumed premises into verified facts. Appendix~\ref{app_prefix} formalizes both readings, measures their divergence, and reports the grounded-context variant.

For each candidate relation $z\in C_k$, the auditor assigns
\begin{equation}
\tau(B_k,z)=
\begin{cases}
\textsc{no}, & \neg\operatorname{SAT}(B_k),\\
\textsc{co}, & \operatorname{SAT}(B_k)\land
\neg\operatorname{SAT}(B_k\cup\{z\}),\\
\textsc{en}, & \operatorname{SAT}(B_k)\land B_k\models z,\\
\textsc{un}, & \text{otherwise}.
\end{cases}
\label{eq:status}
\end{equation}

For directional queries, the implemented nine-way relation system is treated as exhaustive and mutually exclusive. A directional relation is entailed when every alternative direction is infeasible under the active context. For non-directional relations, entailment is determined by membership in the valid relational closure. Under a feasible context, a claim containing multiple relations is contradicted if at least one relation is contradicted and is entailed only if every relation is entailed. All other evaluable cases are assigned \textsc{unknown}. If the active context is already infeasible, the claim is marked \textsc{not-evaluable}.

An \textsc{unknown} claim is evaluable but neither entailed nor contradicted; a \textsc{not-evaluable} claim inherits an infeasible prefix. Neither contributes positive reliability evidence, and both reduce determinacy. The auditor also records the initiating conflict and a bounded repair cost that removes only generated relation instances while keeping the trusted scene fixed. Appendix~\ref{app_auditor} defines these operations and lists all claim- and trace-level features; Appendix~\ref{sec:app_case_studies} provides illustrative entailment, contradiction, and parsed-but-unknown cases.

The auditor receives only the scene, question, and generated trace. It never accesses reference answers or intermediate-claim labels. An entailed status thus means entailment under the active context and implemented relation system, not correctness of the entire trace.

\subsection{Determinacy Profile}\label{sec:applicability}

Parsing a claim and semantically resolving it are distinct. Let
$a_k=\mathds{1}[C_k\neq\varnothing]$ indicate whether claim $c_k$ is symbolizable, and let $g_k=\mathds{1}[\operatorname{SAT}(B_k)]$ indicate whether its active context is feasible. We define trace-level symbolizability as
\begin{equation}
\pi(r)=\frac{1}{K}\sum_{k=1}^{K}a_k.
\end{equation}

We further define the unknown rate among parsed claims using an explicit zero-coverage convention:
\begin{equation}
\nu(r)=
\begin{cases}
\dfrac{\sum_{k=1}^{K}a_k\mathds{1}[\tau_k=\textsc{un}]}{\sum_{k=1}^{K}a_k}, & \pi(r)>0,\\[6pt]
0, & \pi(r)=0.
\end{cases}
\end{equation}
Here, $\tau_k$ aggregates the relation-level verdicts in Equation~\eqref{eq:status} using the conservative rule above. For $C_k=\varnothing$, we set $\tau_k=\textsc{un}$ by convention. Since then $a_k=0$, unparsed claims affect neither $\nu(r)$ nor $d(r)$.

Semantic determinacy measures the effective semantic resolution of the parsed trace. Specifically, it is the fraction of parsed claims that remain evaluable and receive a definite verdict:
\begin{equation}
d(r)=
\begin{cases}
\dfrac{\sum_{k=1}^{K}a_kg_k\,
\mathds{1}[\tau_k\in\{\textsc{en},\textsc{co}\}]}{\sum_{k=1}^{K}a_k},
& \pi(r)>0,\\[6pt]
0, & \pi(r)=0,
\end{cases}
\label{eq:determinacy}
\end{equation}
Claims marked $\textsc{no}$ remain in the denominator but not the numerator. They therefore reduce determinacy rather than being treated as contradicted or silently excluded. Consequently, $d(r)$ measures effective executable coverage by capturing both semantic resolution and the loss of coverage caused by an earlier conflict. It does not encode whether the resolved claims are entailed or contradicted.

A trace with no parsed claims is assigned $d(r)=0$, and $\mathds{1}[\pi(r)=0]$ records this boundary regime. A trace can therefore be highly symbolizable but weakly determinate when recognized claims remain unknown or become blocked. Since verdicts use $A_k$, $d(r)$ measures executable coverage rather than groundedness.

We define overall verifier applicability as
\begin{equation}
\delta(r)=\pi(r)d(r).
\label{eq:applicability}
\end{equation}
Because $\pi(r)$ measures the fraction of parsed claims and $d(r)$ measures the determinate fraction among them, $\delta(r)$ is equivalently the fraction of all claims that are both symbolizable and semantically determinate. It is zero when no claim is parsed and decreases when parsed claims remain unresolved or become not evaluable.

The label-free Determinacy Profile is the coverage descriptor
\begin{equation}
\phi(r)=\big[\pi(r),\nu(r),d(r),\delta(r),\mathds{1}[\pi(r){=}0]\big].
\end{equation}
Its coordinates contain neither verdict polarity nor conclusion status, so the profile describes the execution regime rather than serving as a confidence score. DARC uses $\pi(r)$ and $\delta(r)$; the other coordinates support reporting and analysis.

This separation exposes three regimes: high $\pi(r)$ and $d(r)$ provide substantial executable evidence; high $\pi(r)$ but low $d(r)$ indicates recognizable yet unresolved or blocked claims; low $\pi(r)$ places much of the trace outside the symbolic interface. This trace-level variation motivates DARC's instance-level interactions. 

\subsection{Determinacy-Aware Reliability Composition}

\paragraph{Reliability evidence bank.}
Because no base estimator is assumed to be uniformly reliable, we construct a candidate bank spanning three evidence families. (i) \emph{Constraint-based} candidates include a fixed readout of feasibility and semantic-status features, as well as a learned hybrid that combines auditor features with frozen token representations and incorporates a direct residual derived from entailment and contradiction under the implemented rules. (ii) \emph{Representation-based} probes operate on the same cached model features. (iii) \emph{Decoding-based} scores require no additional training and may remain informative when neither the symbolic solver nor a transferred representation probe is reliable. More details are in Appendix~\ref{app_baselines}.

Supervised candidates are trained once on the source data over the frozen backbone. An optional positive-slope Platt mapping preserves score ordering and therefore does not change AUROC. The evidence bank should thus be viewed as a collection of measurements with different applicability conditions rather than as a set of mandatory modules. DARC treats reliability estimation as an evidence-selection and composition problem. It removes candidates that contain no useful validation ranking information and assigns each retained candidate a trace-dependent contribution.

\paragraph{Composition.}
\label{sec:combiner}

Let $s_j(r)$ denote the validation-calibrated score produced by candidate $j$. DARC first screens and orients candidates using the target validation split. A candidate is retained if
$|A_j-0.5|\geq\epsilon$, where $A_j$ is its validation AUROC and $\epsilon=0.03$. If a retained candidate is negatively oriented, its score is replaced by $1-s_j$.
This procedure addresses two common transfer failures using validation labels only. First, including a nearly random score can dilute informative evidence when many candidate signals are combined. Second, a score whose orientation reverses under transfer can cancel useful evidence in a fixed aggregation. Subsequent regularization controls the contribution of each retained candidate to the final-answer reliability estimate.

Let $\widetilde{\bm s}(r)$ contain the retained and consistently oriented candidate scores. We consider three nested feature designs:
\begin{align}
h_{\mathrm{score}}(r) &= [\widetilde{\bm s}(r)],\\
h_{\mathrm{sym}}(r) &= [h_{\mathrm{score}}(r),
\pi(r)\widetilde{\bm s}(r)],\\
h_{\mathrm{det}}(r) &= [h_{\mathrm{sym}}(r),
\delta(r)\widetilde{\bm s}(r)].
\label{eq:darc_features}
\end{align}
The applicability variables appear only through interactions with candidate scores. They can therefore modify the contribution of each retained score according to trace symbolizability and semantic determinacy, but cannot act as independent reliability predictors. Because $h_{\mathrm{det}}$ retains the raw score block, however, these interactions are additive refinements rather than multiplicative gates: they adjust the marginal contribution of each candidate with applicability but do not force it to vanish when $\delta(r)=0$. The design is applicability-aware rather than applicability-gated, so suppression of constraint evidence at low determinacy is a property of the fitted weights rather than a structural guarantee. Appendix~\ref{app_prefix} discusses a strictly gated variant.

Given a selected feature design $h(r)$, the composer is
\begin{equation}
p(r)=\sigma\big(\bm w^{\top}\operatorname{std}(h(r))+b\big),
\label{eq:darc}
\end{equation}
where standardization statistics come from the target validation split. The parameters minimize binary cross-entropy with $\ell_2$ regularization on $\bm w$. Ordinary validation Brier score selects the identity or a strictly increasing affine-logit, temperature, or monotone-beta calibrator. Strict monotonicity preserves ranking and therefore leaves AUROC unchanged.

DARC uses continuous composition because a locally entailed claim may still occur in a trace whose final answer is incorrect. The coefficient of $\delta(r)s_j(r)$ can change candidate $j$'s contribution with verifier applicability, while the raw score features preserve useful evidence when the profile is uninformative.

The three feature designs form a controlled hierarchy. Selecting among them avoids assuming that the richest interaction structure is always preferable, particularly when the labeled validation split is limited. The feature design and $\ell_2$ regularization strength are selected through deterministic $k$-fold cross-validation within the target validation split. Each pair $(\text{design},\lambda)$ is evaluated using mean held-out AUROC, with the larger value of $\lambda$ preferred in a tie. The selected composer is then refitted on the complete validation split and applied once to the test set. Appendix~\ref{app_hyperparameters} reports the complete search space and reference defaults.

This procedure constitutes target-adapted transfer. It updates neither the language model generator nor any base estimator. Appendix~\ref{app_transfer} provides the complete algorithm and leakage-control procedures.

\section{Evaluation}\label{sec:evaluation}

\begin{table*}[t]
\centering
\small

\setlength{\tabcolsep}{6pt}
\begin{tabular}{ll cc cc cc cc cc}
\toprule
& \multirow{2}{*}{\textbf{Method}} 
& \multicolumn{2}{c}{\textbf{StepGame} (Source)} 
& \multicolumn{2}{c}{\textbf{SpaRTQA}  (Target)} 
& \multicolumn{2}{c}{\textbf{SpaRTUN}  (Target)} 
& \multicolumn{2}{c}{\textbf{SpaceNLI}  (Target)} 
& \multicolumn{2}{c}{\textbf{SpaRP}  (Target)} \\
\cmidrule(lr){3-4} \cmidrule(lr){5-6} \cmidrule(lr){7-8} \cmidrule(lr){9-10} \cmidrule(lr){11-12}
& & AUC $\uparrow$ & BS $\downarrow$ & AUC $\uparrow$ & BS $\downarrow$ & AUC $\uparrow$ & BS $\downarrow$ & AUC $\uparrow$ & BS $\downarrow$ & AUC $\uparrow$ & BS $\downarrow$ \\
\midrule

\multirow{5}{*}{\rotatebox{90}{\textit{Unsup.}}}
& Random & 0.524 & \graycell{0.432} & 0.489 & \graycell{0.413} & 0.494 & \graycell{0.431} & 0.496 & \graycell{0.257} & 0.486 & \graycell{0.265} \\
& Perplexity & 0.573 & 0.250 & 0.510 & \graycell{0.250} & 0.511 & \graycell{0.250} & 0.552 & 0.248 & 0.527 & \graycell{0.250} \\
& Token Entropy & 0.507 & \graycell{0.251} & 0.480 & \graycell{0.250} & 0.491 & \graycell{0.250} & 0.489 & \graycell{0.250} & 0.489 & \graycell{0.250} \\
& MCP & 0.420 & 0.256 & 0.472 & \graycell{0.250} & 0.489 & \graycell{0.250} & 0.485 & \graycell{0.250} & 0.474 & \graycell{0.250} \\
& CCP & 0.428 & 0.255 & 0.464 & 0.250 & 0.481 & \graycell{0.250} & 0.485 & \graycell{0.250} & 0.474 & \graycell{0.250} \\
\midrule
\multirow{3}{*}{\rotatebox{90}{\textit{Sampl.}}}
& Semantic Entropy & 0.522 & \graycell{0.329} & 0.495 & \graycell{0.313} & 0.513 & \graycell{0.276} & 0.514 & \graycell{0.275} & 0.501 & \graycell{0.330} \\
& SelfCheckGPT & 0.486 & \graycell{0.317} & 0.507 & \graycell{0.312} & 0.488 & \graycell{0.274} & 0.506 & \graycell{0.263} & 0.490 & \graycell{0.323} \\
& P(True) & 0.476 & \graycell{0.307} & 0.511 & \graycell{0.318} & 0.510 & \graycell{0.277} & 0.512 & \graycell{0.263} & 0.496 & \graycell{0.322} \\
\midrule
\multirow{4}{*}{\rotatebox{90}{\textit{Neural}}}
& Factoscope & 0.697 & 0.222 & 0.494 & \graycell{0.250} & 0.342 & 0.250 & 0.525 & \graycell{0.250} & 0.494 & \graycell{0.250} \\
& UHead & 0.670 & 0.234 & 0.506 & \graycell{0.250} & 0.356 & 0.250 & 0.508 & \graycell{0.250} & 0.505 & \graycell{0.250} \\
& Neural-Seq & 0.754 & 0.205 & 0.347 & 0.250 & 0.380 & 0.250 & \underline{0.724} & \underline{0.212} & 0.507 & \graycell{0.251} \\
& MLP & 0.508 & \graycell{0.250} & \underline{0.568} & \underline{0.247} & 0.511 & \graycell{0.250} & 0.532 & 0.249 & 0.528 & \graycell{0.250} \\
\midrule
\multirow{3}{*}{\rotatebox{90}{\textit{Symb.}}}
& Constraint-Rule & 0.769 & \underline{0.186} & 0.492 & \graycell{0.250} & 0.449 & 0.250 & 0.496 & \graycell{0.238} & 0.678 & \underline{0.208} \\
& Constraint-Only & 0.767 & 0.207 & 0.499 & \graycell{0.250} & \underline{0.558} & \underline{0.247} & 0.662 & 0.227 & \underline{0.701} & 0.212 \\
& Constraint & \underline{0.790} & 0.196 & 0.504 & \graycell{0.250} & 0.497 & \graycell{0.250} & 0.605 & 0.236 & 0.617 & 0.290 \\
\midrule
\rowcolor{gray!15}
\cellcolor{white} & \textbf{\m} & \textbf{0.844} & \textbf{0.164} & \textbf{0.657} & \textbf{0.230} & \textbf{0.678} & \textbf{0.225} & \textbf{0.792} & \textbf{0.187} & \textbf{0.741} & \textbf{0.188} \\
\bottomrule
\end{tabular}
\caption{Overall performance for final-answer reliability estimation as \textbf{AUROC} (AUC$\uparrow$) and \textbf{class-balanced Brier loss} (BS$\downarrow$) on \textbf{Mistral-7B-Instruct}. Every method uses the identical validation-fit calibration protocol and is scored against the same greedy trace; sampling baselines draw $K{=}10$ additional decodes. Best results are \textbf{bold}; second best are \underline{underlined}. \graycell{Gray} cells mark non-discriminative scores (AUROC${\approx}0.5$) whose calibrated predictions collapse toward a constant.}
\label{tab:result}
\end{table*}

\begin{table*}[t]
\centering
\small
\setlength{\tabcolsep}{5pt}
\begin{tabular}{l cc cc cc cc cc}
\toprule
\multirow{2}{*}{\textbf{Combiner variant}} & \multicolumn{2}{c}{\textbf{StepGame}} & \multicolumn{2}{c}{\textbf{SpaRTQA}} & \multicolumn{2}{c}{\textbf{SpaRTUN}} & \multicolumn{2}{c}{\textbf{SpaceNLI}} & \multicolumn{2}{c}{\textbf{SpaRP}} \\
\cmidrule(lr){2-3} \cmidrule(lr){4-5} \cmidrule(lr){6-7} \cmidrule(lr){8-9} \cmidrule(lr){10-11}
& AUC $\uparrow$ & BS $\downarrow$ & AUC $\uparrow$ & BS $\downarrow$ & AUC $\uparrow$ & BS $\downarrow$ & AUC $\uparrow$ & BS $\downarrow$ & AUC $\uparrow$ & BS $\downarrow$ \\
\midrule
Constraint scorer (pure) & 0.790 & 0.196 & 0.504 & 0.250 & 0.497 & 0.250 & 0.605 & 0.236 & 0.617 & 0.290 \\
MLP probe (pure) & 0.508 & 0.250 & 0.568 & 0.247 & 0.511 & 0.250 & 0.532 & 0.249 & 0.528 & 0.250 \\
Fixed average: Constraint/MLP & 0.769 & 0.206 & 0.568 & 0.247 & 0.511 & 0.250 & 0.608 & 0.236 & 0.612 & 0.246 \\
Oracle choice: Constraint/Neural-Seq$^{\dagger}$ & 0.790 & 0.196 & 0.504 & 0.250 & 0.497 & 0.250 & 0.724 & 0.212 & 0.617 & 0.290 \\
\midrule
Scores-only stacking (forced $h_{\mathrm{score}}$) & 0.811 & 0.178 & 0.617 & 0.235 & 0.642 & 0.229 & 0.760 & 0.196 & 0.689 & 0.208 \\
\quad + Forced $h_{\mathrm{det}}$ with shuffled $\delta(r)$ & 0.814 & 0.179 & 0.622 & 0.237 & 0.655 & 0.235 & 0.763 & 0.196 & 0.690 & 0.207 \\
\quad + Symbolizability interactions (forced $h_{\mathrm{sym}}$) & 0.818 & 0.181 & 0.634 & 0.235 & 0.651 & 0.230 & 0.760 & 0.195 & 0.694 & 0.207 \\
\quad + Determinacy interactions (forced $h_{\mathrm{det}}$) & 0.828 & 0.171 & 0.628 & 0.236 & 0.658 & \underline{0.228} & 0.779 & 0.190 & 0.727 & \textbf{0.188} \\
\midrule
\m w/ evaluable-only determinacy & 0.835 & 0.168 & \underline{0.651} & 0.233 & 0.664 & 0.230 & 0.774 & 0.191 & \underline{0.733} & 0.194 \\
\m w/ grounded context $T_k$
& \underline{0.837} & \underline{0.166}
& 0.648 & \underline{0.232}
& \underline{0.670} & \underline{0.228}
& \underline{0.784} & \underline{0.189}
& 0.730 & \underline{0.192} \\

\rowcolor{gray!15}
\textbf{\m} & \textbf{0.844} & \textbf{0.164} & \textbf{0.657} & \textbf{0.230} & \textbf{0.678} & \textbf{0.225} & \textbf{0.792} & \textbf{0.187} & \textbf{0.741} & \textbf{0.188} \\
Improvement compared to Scores-only stacking & \multicolumn{2}{c}{$+.033^{*}$} & \multicolumn{2}{c}{$+.040^{*}$} & \multicolumn{2}{c}{$+.036^{*}$} & \multicolumn{2}{c}{$+.032^{*}$} & \multicolumn{2}{c}{$+.052^{*}$} \\
\bottomrule
\end{tabular}
\caption{Component ablation on Mistral-7B. Stacked variants share the screened score bank; the determinacy design contains both $\pi(r)$ and $\delta(r)$ interactions. Full \m selects its design, regularization, and calibration on validation. Bold and underline mark the best and second-best deployable results. $^{\dagger}$ denotes a test oracle excluded from highlighting, and $^{*}$ indicates $p<0.05$ versus scores-only stacking.}
\label{tab:ablation}
\end{table*}

We evaluate \m via the following research questions:

\begin{itemize}[leftmargin=*,nosep]
\item \textbf{RQ 1 (Performance):} How does \m compare with existing final-answer reliability estimators?
\item \textbf{RQ 2 (Ablation Study):} Which design choices account for \m's gains beyond scores-only target adaptation?
\item \textbf{RQ 3 (Applicability):} Does semantic determinacy characterize verifier applicability better than parse coverage?
\item \textbf{RQ 4 (Faithfulness):} Are the auditor's local verdicts reliable?
\item \textbf{RQ 5 (Adaptation):} How much target supervision does \m need?
\item \textbf{RQ 6 (Efficiency):} Is \m computationally efficient?
\end{itemize}

\subsection{Experimental Setup}

\subsubsection{Datasets}
We evaluate our method on five datasets. Supervised base scorers are trained on StepGame and transferred to four target datasets, namely SpaRTQA, SpaRTUN, SpaceNLI, and SpaRP, which span diverse spatial relations and answer formats. For each target dataset, its validation split is used to fit DARC and the post-hoc calibrator. (More details in Appendix~\ref{app_datasets}).

\subsubsection{Models and Labels}
Four frozen backbone models generate the reasoning traces used in our evaluation. Final-answer correctness is determined by a deterministic match between the parsed final answer and the benchmark answer. A separate Qwen3-1.7B judge labels only the intermediate reasoning claims used to train supervised base scorers. These labels show strong agreement with a human annotator and three larger reference judges, as detailed in Appendix~\ref{app_models}. The auditor does not access any of these labels.

\subsubsection{Baselines}
We compare \m against training-free decoding scores, sampling-based estimators, neural probes, and constraint-based scorers. Each baseline uses only the information available to its method family. Decoding-based methods rely on token probabilities, sampling-based methods use $K$ stochastic generations, neural probes operate on cached model activations, and constraint-based methods use generated text and auditor-derived features. \m integrates the resulting scalar scores with the Determinacy Profile. Additional details are provided in Appendix~\ref{app_baselines}.

\subsubsection{Metrics}
We report AUROC to evaluate trace ranking and class-balanced Brier loss to measure probability estimation error. More details are in Appendix~\ref{app_metrics}.
The main evaluation uses structured claims for controlled claim--token alignment. On automatically segmented free-form chain-of-thought traces, \m retains AUROC gains of $0.030$--$0.040$ over scores-only stacking (Appendix~\ref{app_generation}).
Appendices~\ref{sec:app_prompt_templates} and~\ref{app_reproducibility} provide the prompt contracts and reproducibility controls, respectively.

\subsection{RQ 1: Overall Performance}

Table~\ref{tab:result} shows that \m has the best AUROC and class-balanced Brier loss on all five datasets with Mistral-7B. It improves over scores-only stacking in all 20 backbone--dataset settings, significantly in 19 (Appendix~\ref{app_significance}); Mistral-7B results also favor \m across ordinary Brier, NLL, ECE, and AURC, with isolated ECE exceptions. Complete results for the other backbones are in Appendix~\ref{app_additional}. On Mistral-7B, sampling estimators use $K{=}10$ extra generations yet remain near chance. Gray cells mark non-discriminative scores whose calibration collapses toward a constant.

No evidence family is uniformly strongest, as constraint-based, neural, and decoding-based estimators fail on different targets. By using target-validation data to retain, orient, and condition complementary scores, \textbf{\m provides the strongest overall final-answer reliability estimates}. 

\subsection{RQ 2: Ablation Study}

Applicability-aware composition, rather than generic target adaptation alone, accounts for \m's gains. In Table~\ref{tab:ablation}, the first two rows use one scorer, the fixed average equally weights Constraint and MLP, and the restricted oracle chooses between Constraint and Neural-Seq by test AUROC. All learned variants use the same screened bank and validation protocol: scores only, symbolizability, and determinacy force $h_{\mathrm{score}}$, $h_{\mathrm{sym}}$, and $h_{\mathrm{det}}$, respectively; the shuffled row also uses $h_{\mathrm{det}}$ but permutes $\delta(r)$ across traces. Full \m selects the design on validation. Scores-only stacking remains $0.032$--$0.052$ AUROC below full \m; shuffled determinacy adds little, whereas trace-aligned interactions improve consistently.

The evaluable-only variant excludes prefix-blocked claims from determinacy; it remains $0.014$--$0.044$ above scores-only stacking, while including blocked claims adds another $0.006$--$0.018$. The grounded-context variant recomputes auditor features under $T_k$ and retains about $78\%$ of the full gain, so assumption propagation is not the sole explanation. Full \m attains the best or tied-best deployable AUROC and Brier loss throughout, with significant AUROC gains on all five datasets; Appendix~\ref{app_significance} reports all backbones. Thus, \textbf{RQ 2 shows that trace-aligned applicability interactions, including the treatment of blocked claims, provide gains beyond generic adaptation}.

\begin{figure}[t]
    \centering
    \includegraphics[width=0.48\linewidth]{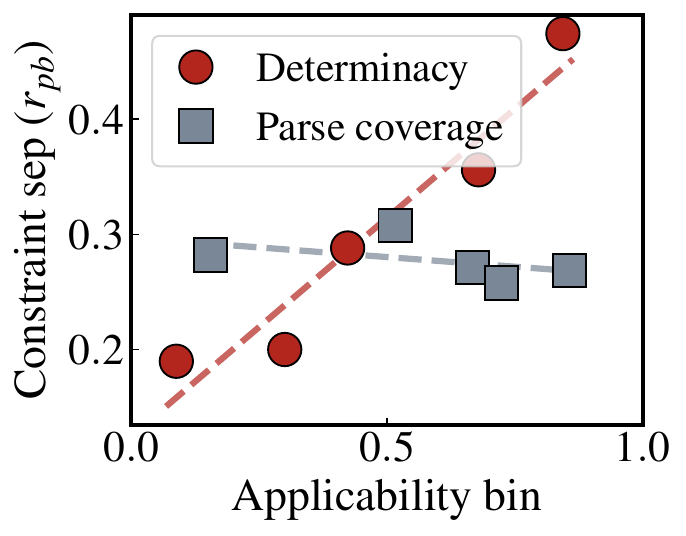}
    \hfill
    \includegraphics[width=0.48\linewidth]{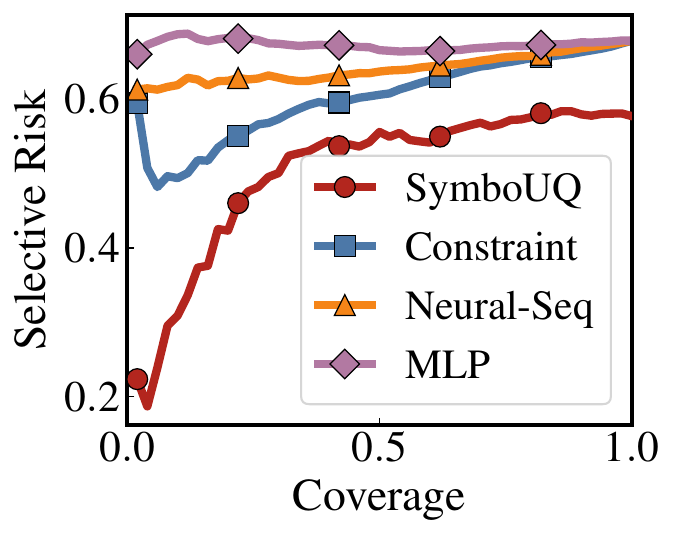}
    \caption{Trace-level applicability and selective final-answer reliability. Left: constraint-score separation across within-dataset bins of determinacy and parse coverage; higher is more informative. Right: risk--coverage on four transfer benchmarks; lower risk is better. }
    \label{fig:analysis}
\end{figure}

\subsection{RQ 3: Determinacy Characterizes Applicability}
Figure~\ref{fig:analysis} shows that semantic determinacy characterizes verifier applicability better than parse coverage. Across within-dataset bins, constraint-score separation increases nearly monotonically with determinacy but not with parse coverage. Consistently, \m achieves the lowest selective risk in the high-confidence regime across the transfer benchmarks.

These results explain the cross-domain variation detailed in Appendix~\ref{app_constraint_utility}. Parse coverage measures access to the verifier, whereas determinacy measures whether execution yields a definite verdict. DARC therefore augments the symbolizability interactions based on $\pi(r)$ with applicability interactions based on $\delta(r)=\pi(r)d(r)$, rather than relying on parse success alone. Although the analysis is associative, \textbf{RQ 3 shows that semantic determinacy characterizes verifier applicability more faithfully than parse coverage.}

\subsection{RQ 4--RQ 6: Auditor Faithfulness, Adaptation, and Efficiency}

The auditor provides useful but imperfect local evidence, with entailed precision of $0.81$--$0.87$ and contradicted precision of $0.64$--$0.74$ (Appendix~\ref{app_auditor_eval}); it also distinguishes all $19{,}234$ controlled relation pairs (Appendix~\ref{app_counterfactual}). Useful transfer emerges around 128 target labels (Appendix~\ref{app_adaptation}). The reported two-score cached-feature path contains 8.13M trainable parameters, trains in 9.0 minutes, and processes 159.3 traces/s (Appendix~\ref{app_efficiency}). Thus, \textbf{RQ 4--RQ 6 establish qualified faithfulness, modest supervision, and competitive cached-feature efficiency.}

\section{Related Work}\label{sec:related-work}

\subsection{Spatial Reasoning in LLMs}

Spatial reasoning requires models to maintain coherent representations of entities and relations across multiple inference steps. However, LLMs become increasingly unreliable as relational complexity grows and often rely on shallow patterns that generalize poorly across distributions \citep{rizvi-etal-2024-sparp,khalid-etal-2025-large}. Prior work improves spatial reasoning by introducing structured intermediate representations, including symbolic reasoning chains, coordinate-based representations, and visualized spatial layouts, to reduce ambiguity in free-form reasoning \citep{hu-etal-2024-chain,cho-etal-2025-power}. These methods primarily aim to improve final-answer accuracy. In contrast, we investigate whether the spatial structure expressed in a generated reasoning trace can provide evidence about whether its final answer is correct. We preserve claim order to capture how spatial constraints accumulate and how an earlier inconsistency affects the evaluability of subsequent claims.

\subsection{Uncertainty Quantification and Verification}

Existing UQ methods derive reliability signals from verbalized confidence \citep{kadavath-etal-2022-know-what,yona-etal-2024-intrinsic-uncertainty}, token-level statistics \citep{fadeeva-etal-2024-fact}, repeated generations \citep{manakul-etal-2023-selfcheckgpt}, semantic agreement \citep{kuhn-etal-2023-semantic-uncertainty,kossen-etal-2024-semantic-entropy-probes}, and internal representations \citep{azaria-mitchell-2023-internal-state,he-etal-2024-llm,shelmanov-etal-2025-head}. Process supervision and learned step-level verifiers extend reliability assessment to intermediate reasoning steps \citep{cobbe-etal-2021-training-verifiers,lightman-etal-2023-verify-step}. However, these statistical and learned signals do not necessarily execute the domain semantics underlying each generated claim.
Formal verification provides a complementary source of evidence by explicitly checking generated outputs against predefined rules. Existing approaches verify safety properties \citep{suresh-etal-2025-beaver}, symbolic causal derivations \citep{he-etal-2026-doverifier}, and structured reasoning through analytical constraints or penalties \citep{manchingal-etal-2026-energy}. Such verification can provide precise local evidence within its supported formal language, but its applicability to free-form reasoning is inherently partial. 
Our work focuses on this gap between access to a verifier and the informativeness of its output. Rather than treating every parsed claim as verified, \m distinguishes whether a claim can be represented in the verifier's language from whether its execution yields a definite semantic verdict. It then uses this trace-dependent applicability to condition the composition of constraint-based, representation-based, and decoding-based reliability signals.

\section{Conclusion}
In this paper, we propose \m, a symbolic UQ framework that estimates final-answer reliability from only one spatial reasoning trace by separating textual symbolizability from semantic determinacy. It includes three key components: a Layout Auditor, a label-free Determinacy Profile, and DARC that composes constraint-based, representation-based, and decoding-based evidence through applicability-conditioned
interactions to estimate final-answer correctness. Extensive experiments on five benchmarks and four frozen LLM backbones show that \m improves AUROC over scores-only stacking in every setting and achieves, on average, approximately $8\%$ higher AUROC and $7\%$ lower class-balanced Brier loss than the strongest external baseline in each setting. Ablation studies demonstrate improvements beyond generic target adaptation, while trace-level analysis shows that semantic determinacy characterizes constraint utility more faithfully than parse coverage.


\clearpage
\newpage
\bibliography{aaai2027}


\clearpage
\newpage
\appendix
\setcounter{secnumdepth}{2}
\section*{Technical Appendix}

\section{Illustrative Example}\label{app:illustrative_example}

This section follows one trace through the complete \m pipeline. The example is constructed to expose a failure mode that parse coverage alone cannot identify: almost every claim can be translated into a valid spatial relation, yet one contradicted claim makes the remaining prefix infeasible. The Layout Auditor localizes this transition, the Determinacy Profile quantifies the resulting loss of usable symbolic evidence, and DARC composes that evidence with complementary neural and decoding signals.

\subsection{Input and Generated Trace}

Consider the following trusted scene:
\begin{quote}\small
$A$ is above $B$. $B$ is right of $C$. $D$ is near $E$.
\end{quote}
The question asks for the position of $A$ relative to $C$. The first two scene relations jointly imply that $A$ is upper right of $C$. Suppose that the frozen generator instead produces the following ordered trace:
\begin{enumerate}[leftmargin=*,nosep]
    \item $B$ is right of $C$.
    \item $A$ is upper right of $C$.
    \item $D$ is near $C$.
    \item The arrangement is now clear.
    \item $A$ is upper left of $C$.
    \item \textbf{Conclusion:} $A$ is upper left of $C$.
\end{enumerate}
The conclusion is incorrect, but the trace is deliberately challenging for output-only uncertainty estimation. It is fluent, it contains a valid intermediate deduction in Claim~2, and five of its six claims use relations supported by the auditor.

\subsection{Sequential Layout Audit}

The auditor first normalizes the scene into
\begin{equation}
D=\{(A,\textsc{above},B),(B,\textsc{right},C),
(D,\textsc{near},E)\}.
\end{equation}
It then parses and executes each claim in order. Table~\ref{tab:illustrative_audit} shows the resulting state. The active context for Claim~$k$ is the trusted scene together with all earlier symbolizable claims, as defined by $A_k$ in Equation~\eqref{eq:prefix}.

\begin{table*}[t]
\centering
\small
\setlength{\tabcolsep}{12pt}
\begin{tabular}{c p{0.20\textwidth} c c p{0.43\textwidth}}
\toprule
$k$ & Generated claim & Parsed & Verdict & Auditor rationale \\
\midrule
1 & $B$ is right of $C$. & yes & \textsc{en} &
The claim directly restates a trusted scene relation. \\
2 & $A$ is upper right of $C$. & yes & \textsc{en} &
Above$(A,B)$ and right$(B,C)$ jointly constrain $A$ to the upper right of $C$. \\
3 & $D$ is near $C$. & yes & \textsc{un} &
Both entities are grounded, but near$(D,E)$ provides neither support nor a contradiction because \textsc{near} is not transitive. \\
4 & The arrangement is now clear. & no & \textsc{un} &
No supported relation is extracted. The claim contributes no positive symbolic evidence. \\
5 & $A$ is upper left of $C$. & yes & \textsc{co} &
The horizontal component conflicts with the trusted implication that $A$ is right of $C$. Adding the claim makes the constraint system infeasible. \\
6 & Conclusion: $A$ is upper left of $C$. & yes & \textsc{no} &
Claim~5 has entered the assumed prefix, so $A_6$ is already infeasible. The conclusion is therefore not evaluable rather than counted as another independent contradiction. \\
\bottomrule
\end{tabular}
\caption{End-to-end Layout Auditor execution for the illustrative trace. The statuses are entailed (\textsc{en}), contradicted (\textsc{co}), unknown (\textsc{un}), and not evaluable (\textsc{no}).}
\label{tab:illustrative_audit}
\end{table*}

This execution makes three distinctions that a binary consistency check would miss. First, Claim~3 is parsed but unresolved, so recognizing its syntax does not make it evidence of correctness. Second, Claim~5 is the first localized conflict. Because its contradiction follows from the fixed scene, removing earlier generated relations cannot repair it within the auditor's generated-premise repair search. Third, Claim~6 receives \textsc{no}, not \textsc{co}: once the prefix is infeasible, the solver cannot attribute a fresh semantic verdict to the conclusion. This prevents one error from being counted repeatedly as several independent contradictions while still recording the coverage lost after the conflict.

\subsection{Determinacy Profile}

There are $K=6$ claims, of which five are symbolizable. The symbolizability of the trace is therefore
\begin{equation}
\pi(r^-)=\frac{5}{6}.
\end{equation}
Among the five parsed claims, Claim~3 is unknown, so $\nu(r^-)=1/5$. Claims~1, 2, and 5 receive definite verdicts under feasible active contexts. Claim~6 remains in the parsed-claim denominator but not in the determinacy numerator because its prefix is infeasible. Consequently,
\begin{equation}
d(r^-)=\frac{3}{5},
\qquad
\delta(r^-)=\pi(r^-)d(r^-)=\frac{1}{2}.
\end{equation}
Parse coverage alone would report a seemingly strong value of $5/6$. In contrast, $\delta(r^-)=1/2$ states the operational fact needed by the reliability layer: only half of all claims are both symbolizable and semantically determinate. The full label-free profile is
\begin{equation}
\phi(r^-)=\left[\frac{5}{6},\frac{1}{5},
\frac{3}{5},\frac{1}{2},0\right].
\end{equation}
Verdict polarity is intentionally absent from this profile. The contradiction, infeasible-prefix transition, conclusion status, and repair evidence remain in the constraint-based scores, while $\phi(r^-)$ describes how much of the trace those scores could meaningfully audit.

\subsection{Reliability Composition}

The same trace is also processed by the frozen representation-based and decoding-based estimators. DARC does not assume that any one evidence family is reliable on every target. Using only the labeled target-validation split, it removes candidates whose ranking is near random, reverses negatively oriented retained scores, and forms the vector $\widetilde{\bm s}(r^-)$. For this example, the final feature vector is
\begin{equation}
\begin{aligned}
h_{\mathrm{det}}(r^-)
=\big[
&\widetilde{\bm s}(r^-),\;
\tfrac{5}{6}\widetilde{\bm s}(r^-),\;
\tfrac{1}{2}\widetilde{\bm s}(r^-)
\big].
\end{aligned}
\end{equation}
The raw block preserves complementary evidence even though the symbolic context becomes incomplete. The second block lets validation fitting adjust each score according to whether the trace is expressible in the auditor's language. The third block separately adjusts it according to effective semantic resolution. After validation-only standardization, DARC applies Equation~\eqref{eq:darc}, and the selected post-hoc calibrator maps the result to the final-answer reliability estimate $p(r^-)$. No reference answer or test label is used in auditing, feature construction, candidate selection, fitting, or calibration for this test trace. The benchmark answer is consulted only afterward to evaluate the already produced estimate.

\subsection{Why Determinacy Adds Information}

To isolate the role of determinacy, consider a minimally corrected trace $r^+$ that replaces ``upper left'' with ``upper right'' in Claims~5 and 6 while leaving every other token-level claim slot unchanged. Claims~5 and 6 then both become entailed. The corrected and erroneous traces have identical symbolizability,
\begin{equation}
\pi(r^+)=\pi(r^-)=\frac{5}{6},
\end{equation}
so a parse-coverage gate cannot distinguish them. Their executable coverage is different:
\begin{equation}
d(r^+)=\frac{4}{5},\quad
\delta(r^+)=\frac{2}{3},
\qquad
d(r^-)=\frac{3}{5},\quad
\delta(r^-)=\frac{1}{2}.
\end{equation}
The auditor also changes from two final entailments in $r^+$ to a contradiction followed by a blocked conclusion in $r^-$. Thus, \m exposes both \emph{what} semantic evidence was observed through the constraint scores and \emph{how applicable} that evidence was through the Determinacy Profile. DARC can then learn, from target-validation data, how strongly each evidence source should contribute in these two regimes instead of treating their identical parse rates as equally informative.

This controlled pair illustrates the complete argument of the method: symbolizability measures access to the formal interface, semantic determinacy measures the usable outcome of execution, and applicability-aware composition combines that partial evidence without discarding non-symbolic signals. The example establishes the mechanism rather than serving as standalone empirical proof. The trace-level trends in Figure~\ref{fig:analysis} and the controlled composition ablations in Table~\ref{tab:ablation} provide the corresponding empirical evidence that determinacy is more informative than parse coverage and that its aligned interactions improve reliability estimation.

\section{Auditor Implementation}\label{app_auditor}

\subsection{Canonical Relation System}

The Layout Auditor recognizes seven relation families. Directional relations are above, below, left, right, the four diagonals, and \textit{same-position}. The last label is the ninth class in directional benchmarks and is distinct from topological \textit{overlap}. Containment includes inside, contains, and their negations. Distance contains near and far. Topology contains touching, disconnected, and overlap. Location represents the current position of an entity. Path relations represent start and end locations. All relation aliases are normalized before solving.

Inverse rules are applied to directional and containment relations. Symmetry is applied only where valid. Transitivity is restricted to inside and contains. Near and touching are deliberately not treated as transitive. These choices prevent the solver from converting linguistic similarity into unsupported spatial closure.

The directional solver creates two independent systems of difference constraints. Feasibility requires both axes to be satisfiable. The symbolic solver computes closure until no new fact can be added, then tests incompatible relation pairs. The combined result is exact relative to these implemented rules and their supported language; it is not a claim of completeness for unrestricted natural-language spatial reasoning.

\subsection{Claim Status and Repair}

The status of a claim with several extracted relations is aggregated conservatively. One contradicted relation makes the claim contradicted. Every relation must be entailed for the claim to be entailed. A mixture of entailed and unresolved relations is unknown. Text with no extracted relation is also unknown, but it is excluded from rates that are explicitly conditioned on parsing.

For a contradicted relation $z$, let $P_k=A_k\setminus D$ be the multiset of earlier generated relation instances. The context-preserving repair cost is
\begin{equation}
R_k(z)=\min_{S\subseteq P_k}|S|
\quad\text{s.t.}\quad
\operatorname{SAT}\big(D\cup(P_k\setminus S)\cup\{z\}\big).
\label{eq:repair}
\end{equation}
Repair search preserves every scene relation and enumerates subsets of one to three generated instances. A value of four denotes that the claim could not be repaired within this budget. This bounded value is a feature, not the exact repair distance when the true distance exceeds three. For claims that are not contradicted, the corresponding repair feature is set to zero by convention.

The auditor produces 23 claim features and 16 trace features. Claim features describe parsing, grounding, feasibility, status, first conflict, repair, position, claim type, and relation family. Trace features summarize context size, claim count, parse rate, grounding, full feasibility, status rates, first conflict position, repair, and conclusion status. The cache rebuild script recomputes these features from deployment-visible text and verifies their dimensions before training.

\subsection{Assumed and Grounded Contexts}\label{app_prefix}

The auditor evaluates claim $k$ against the assumed context $A_k$ of Equation~\eqref{eq:prefix}, which admits every earlier symbolizable claim as a working hypothesis, rather than the grounded context $T_k$ of Equation~\eqref{eq:trusted}. In grounded replay, $\tau_j^T$ denotes claim $j$'s status computed recursively under $T_j$, so only grounded entailments enter later grounded contexts. This subsection makes the consequences explicit and measures them.

Under $A_k$, a verdict of \textsc{entailed} is a conditional statement: the claim follows from the trusted scene together with the hypotheses the trace has already introduced. It is therefore possible for a chain of ungrounded but mutually consistent claims to accumulate conditional entailments. If the scene relates neither $A$ to $B$ nor $B$ to $C$, then a trace asserting that $A$ is left of $B$ and that $B$ is left of $C$ receives \textsc{unknown} for both claims, and a third claim that $A$ is left of $C$ is then \textsc{entailed} under $A_3$ even though it is \textsc{unknown} under $T_3$.

Two properties of the auditor limit how such a verdict is reported. The unresolved premises contribute nothing to the entailment or contradiction rates, and they lower semantic determinacy $d(r)$ and hence applicability $\delta(r)$. Neither property removes the underlying degeneracy, and we state its scope precisely rather than claim that it is prevented.

First, determinacy is a coverage ratio whose numerator counts verdicts obtained under $A_k$. Extending the example to $q$ unresolved premises followed by $m$ conditionally entailed consequences gives
\begin{equation}
d(r)=\frac{m}{m+q}\xrightarrow[m\to\infty]{}1 ,
\label{eq:det_dilution}
\end{equation}
so a long derivation resting on a few unsupported premises can be reported as highly determinate. The penalty applied to the $q$ unresolved claims is diluted rather than propagated to their consequences, because the auditor records a verdict per claim and does not maintain the provenance that would mark a consequence as assumption-dependent when features are constructed.

Second, the $\delta(r)$ interactions in Equation~\eqref{eq:darc_features} are additive terms alongside the raw score block, not multiplicative gates, so low applicability reduces the marginal weight of constraint evidence without bounding it. Whether constraint evidence is in fact suppressed at low determinacy is therefore a property of the fitted weights, not of the feature design.

The conditional entailment above is consequently recorded as coherence of the generated chain rather than as grounded support, and $d(r)$ bounds executable coverage rather than groundedness. The measurements below bound how often the two readings diverge in practice, which is the sense in which the reported results rest predominantly on grounded evidence.

We retain $A_k$ rather than $T_k$ for two reasons. First, the auditor supplies evidence for final-answer reliability rather than serving as a proof checker: the observation that a trace contradicts its own earlier commitments can be diagnostic of an incorrect final answer, and this observation is only available when generated hypotheses remain in the context. Second, error propagation, first-conflict position, and repair cost are defined over the generated chain and become vacuous under $T_k$, because an unverified claim that causes a later conflict would simply have been discarded before the conflict could be observed.

Table~\ref{tab:prefix} quantifies the difference. For each backbone and dataset, we replay the cached parsed relations of $250$ test traces under both context definitions and compare the resulting verdicts. Overall, only $2.4\%$ of \textsc{entailed} verdicts are assumption-dependent, meaning that they do not survive when unverified premises are withheld; the dataset-level rate ranges from $1.2\%$ on StepGame to $5.8\%$ on SpaceNLI. The aggregate entailment rate is nearly unchanged under grounded-context replay ($0.327$ under $T_k$ versus $0.324$ under $A_k$) and is sometimes higher, because withholding an erroneous hypothesis can restore feasibility and allow a later claim to be resolved. Contradiction evidence is more context-sensitive: $12.9\%$ of \textsc{contradicted} verdicts are assumption-dependent overall, with dataset-level rates ranging from $9.4\%$ on StepGame to $28.7\%$ on SpaceNLI. The grounded context also yields a higher aggregate contradiction rate ($0.095$ versus $0.073$), indicating that unsupported hypotheses can block or alter later contradiction judgments. This greater sensitivity is consistent with the lower precision of contradicted verdicts reported in Table~\ref{tab:auditor}. These results show that the auditor's entailment evidence is predominantly grounded, whereas the contradiction channel more often reflects coherence with the generated hypothesis chain.

\begin{table}[t]
\centering
\small
\setlength{\tabcolsep}{4pt}
\resizebox{\columnwidth}{!}{%
\begin{tabular}{l c cc c cc c}
\toprule
& & \multicolumn{3}{c}{\textsc{entailed}} & \multicolumn{3}{c}{\textsc{contradicted}} \\
\cmidrule(lr){3-5}\cmidrule(lr){6-8}
\textbf{Dataset} & Claims & $A_k$ & $T_k$ & Dep. & $A_k$ & $T_k$ & Dep. \\
\midrule
StepGame & 4{,}147
& 0.472 & 0.481 & 0.012
& 0.112 & 0.146 & 0.094 \\

SpaRTQA & 4{,}008
& 0.168 & 0.171 & 0.031
& 0.052 & 0.067 & 0.183 \\

SpaRTUN & 2{,}854
& 0.214 & 0.207 & 0.047
& 0.014 & 0.018 & 0.261 \\

SpaceNLI & 1{,}747
& 0.281 & 0.269 & 0.058
& 0.034 & 0.039 & 0.287 \\

SpaRP & 5{,}207
& 0.402 & 0.411 & 0.019
& 0.104 & 0.138 & 0.112 \\
\midrule
All & 17{,}963
& 0.324 & 0.327 & 0.024
& 0.073 & 0.095 & 0.129 \\
\bottomrule
\end{tabular}
}
\caption{Context semantics of the Layout Auditor, pooled over four
backbones ($250$ test traces per backbone and dataset, $17{,}963$
parsed claims). Columns $A_k$ and $T_k$ give the verdict rate among
parsed claims under the assumed context of
Equation~\eqref{eq:prefix} and the grounded context of
Equation~\eqref{eq:trusted}, respectively. The grounded-context replay
uses the same parsed relations but admits only previously entailed
claims. Dep.\ is the fraction of verdicts produced under $A_k$ that
depend on at least one unverified prior claim, namely, verdicts that do
not survive under $T_k$. Entailment evidence is predominantly grounded:
only $2.4\%$ of entailed verdicts are assumption-dependent overall, with
dataset-level rates ranging from $1.2\%$ to $5.8\%$. Contradiction
evidence is more sensitive to the generated hypothesis chain:
$12.9\%$ of contradicted verdicts are assumption-dependent overall,
with dataset-level rates ranging from $9.4\%$ to $28.7\%$. The
grounded-context replay produces nearly unchanged entailment rates
($0.327$ versus $0.324$ overall) but a higher contradiction rate
($0.095$ versus $0.073$), consistent with unsupported hypotheses
occasionally blocking or altering later contradiction judgments. This
greater context sensitivity is also consistent with the lower precision
of contradicted verdicts reported in Table~\ref{tab:auditor}. All
quantities are recomputed from cached parsed relations and involve no
additional model inference.}
\label{tab:prefix}
\end{table}

These measurements are claim-level: they quantify how many individual verdicts change when unverified premises are withheld, and they bound the divergence between the two readings on the traces we use. They do not by themselves establish that $A_k$ is the better choice of context semantics. We therefore additionally evaluate a grounded-only auditor under $T_k$ end to end in Table~\ref{tab:ablation}. This variant remains above scores-only stacking on all five datasets and retains approximately $78\%$ of the full AUROC gain, indicating that most of the improvement does not depend on assumption propagation. Three other context designs remain untested here: a non-committal auditor that never admits an unresolved claim while still detecting self-contradiction; a dual-channel feature set exposing grounded and coherence verdicts in parallel; and a multi-world auditor maintaining several feasible completions. A fourth variant concerns the composer rather than the auditor: restricting constraint candidates to enter only through the $\pi(r)$ and $\delta(r)$ interaction terms would make suppression at low applicability structural rather than learned, at the cost of discarding constraint evidence whenever the profile is uninformative. These four remaining variants are compatible with the Determinacy Profile and with the selection procedure of Algorithm~\ref{alg:darc}, and each would need its own validation-selected configuration. Comparing them under the shared protocol, together with provenance tracking that would let a conditional entailment inherit the determinacy penalty of the premises it depends on, is a natural extension of this work.

\section{DARC Protocol}\label{app_transfer}

Algorithm~\ref{alg:darc} gives the complete target adaptation procedure. Candidate screening and orientation use only target validation labels. The threshold $\epsilon$ is fixed before evaluation. The validation partition is deterministic, which makes the complete composition reproducible.

\begin{algorithm}[t]
\caption{Target adaptation of DARC}
\label{alg:darc}
\begin{algorithmic}[1]
\REQUIRE Validation scores $S_v$, labels $y_v$, profiles $\Phi_v$, test scores $S_t$, test profiles $\Phi_t$
\ENSURE Test final-answer reliability scores $p_t$
\STATE Use profiles computed by Eq.~\eqref{eq:determinacy}, where claims marked $\textsc{no}$ reduce determinacy
\STATE If validation contains one class, return the fixed constraint score
\STATE Compute validation AUROC $A_j$ for each candidate $j$
\STATE Retain candidates with $|A_j-0.5|\geq\epsilon$
\STATE If no candidate remains, return the best validation candidate
\STATE Replace score $s_j$ by $1-s_j$ when $A_j<0.5$
\STATE Partition validation indices into $k$ deterministic folds
\STATE If a fold is degenerate, use $h_{\mathrm{det}}$ with reference regularization
\FOR{each feature design $h$ and regularization value $\lambda$}
    \STATE Fit on $k{-}1$ folds and record held-out AUROC; average over folds
\ENDFOR
    \STATE Select the best $(h,\lambda)$ (prefer larger $\lambda$ in a tie) and refit on all validation examples
\STATE Select and fit the identity or a strictly increasing calibrator by ordinary validation Brier
\STATE Apply validation standardization, the frozen composer, and the calibrator to $(S_t,\Phi_t)$
\RETURN $p_t$
\end{algorithmic}
\end{algorithm}

The candidate bank contains six supervised scores and five training-free scores. The supervised set contains the primary Constraint scorer, Constraint-Only, Neural-Seq, MLP, UHead, and Factoscope. The training-free set contains Constraint-Rule, perplexity, token entropy, MCP, and CCP. Random is a reference and is never supplied to DARC.

Candidate screening is part of the reported method. It prevents a large set of nearly random transferred scores from diluting a useful signal. Orientation allows an informative but reversed validation ranking to contribute consistently. Both operations make the method target-adapted rather than zero-shot. If a validation split contains one class, fitting and selection are undefined. The implementation then uses a fixed constraint score as a conservative fallback. If an internal validation partition is degenerate, it uses the determinacy design with reference regularization.

The score, symbolizability, and determinacy feature designs are nested as shown in Eq.~\eqref{eq:darc_features}. The regularization grid is reported in Table~\ref{tab:search-space}. After selection, test labels are not loaded by the composer. Permuting them therefore leaves every predicted score unchanged.

\section{Experimental Setup}\label{app_experiments}

\subsection{Datasets}\label{app_datasets}

StepGame is the source domain and contains directional chains with controlled reasoning depth \citep{shi-etal-2022-stepgame}. SpaRTQA uses spatial questions over richer natural-language scenes \citep{mirzaee-etal-2021-spartqa}. SpaRTUN covers several relation families and set-valued answers \citep{mirzaee-kordjamshidi-2022-transfer}. SpaceNLI asks whether a spatial hypothesis is entailed, contradicted, or neutral \citep{abzianidze-etal-2023-spacenli}. SpaRP re-renders spatial QA with explicit relation-composition paths over a rich relation vocabulary \citep{rizvi-etal-2024-sparp}. Table~\ref{tab:dataset} gives validation and test sizes.

\begin{table}[t]
\centering
\small
\setlength{\tabcolsep}{9pt}
\begin{tabular}{l l c cc}
\toprule
\textbf{Dataset} & \textbf{Answer space} & \textbf{Train} & \textbf{Val} & \textbf{Test} \\
\midrule
StepGame  & 9-way direction & 10000 & 300  & 3000 \\
SpaRTQA   & 4-way relation  & -  & 300  & 3000 \\
SpaRTUN   & relation set    & -  & 300  & 3000 \\
SpaceNLI  & 3-way NLI       & -  & 300  & 3000 \\
SpaRP     & 5-way relation  & -  & 300  & 1500 \\
\bottomrule
\end{tabular}
\caption{Benchmark dataset statistics.}
\label{tab:dataset}
\end{table}

Each target validation split is used for composition and calibration, and its test split is evaluated once. Base scorer parameters are learned only from StepGame. This protocol measures adaptation of uncertainty evidence rather than adaptation of the reasoning model.

\subsection{Backbones and Judge}\label{app_models}

Four frozen backbones generate independent traces and cached features: Mistral-7B-Instruct-v0.3 \citep{jiang2023mistral}, Llama-3.1-8B-Instruct \citep{meta-ai-2024-llama31}, Gemma-2-9B-it \citep{google-2024-gemma2}, and Qwen3-8B \citep{qwen-2025-qwen3}. A separate Qwen3-1.7B model from the same Qwen3 family supplies intermediate verification labels. Table~\ref{tab:models} lists their roles.

\begin{table}[ht]
\centering
\footnotesize
\setlength{\tabcolsep}{18pt}
\begin{tabular}{@{}l cc@{}}
\toprule
\textbf{Model} & \textbf{Params} & \textbf{Role}\\
\midrule
 Mistral-7B-Instruct-v0.3        & 7B  & Backbone\\
 Llama-3.1-8B-Instruct           & 8B  & Backbone\\
 Gemma-2-9B-it                   & 9B  & Backbone\\
 Qwen3-8B                        & 8B  & Backbone\\
 Qwen3-1.7B                      & 1.7B & Judge\\
\bottomrule
\end{tabular}
\caption{Frozen generation backbones and the separate verification model.}
\label{tab:models}
\end{table}

\begin{table*}[t]
\centering\small
\setlength{\tabcolsep}{3.6pt}
\begin{tabular}{l cc cc cc cc cc}
\toprule
\multirow{2.5}{*}{\textbf{Reference Judge}} & \multicolumn{2}{c}{\textbf{StepGame}} & \multicolumn{2}{c}{\textbf{SpaRTQA}} & \multicolumn{2}{c}{\textbf{SpaRTUN}} & \multicolumn{2}{c}{\textbf{SpaceNLI}} & \multicolumn{2}{c}{\textbf{SpaRP}} \\
\cmidrule(lr){2-3} \cmidrule(lr){4-5} \cmidrule(lr){6-7} \cmidrule(lr){8-9} \cmidrule(lr){10-11}
 & Agree.(\%)$\uparrow$ & $\kappa\uparrow$ & Agree.(\%)$\uparrow$ & $\kappa\uparrow$ & Agree.(\%)$\uparrow$ & $\kappa\uparrow$ & Agree.(\%)$\uparrow$ & $\kappa\uparrow$ & Agree.(\%)$\uparrow$ & $\kappa\uparrow$ \\
\midrule
\textbf{Human} \textit{(500 samples, 1426 claims)} & 93.1 & 0.83 & 93.2 & 0.86 & 95.9 & 0.89 & 97.6 & 0.94 & 94.2 & 0.85 \\
\midrule
\multicolumn{11}{l}{\textit{LLM Judges (Full evaluation pool)}} \\
Mistral-Small-3.2-24B-Instruct-2506 & 93.5 & 0.84 & 93.7 & 0.80 & 96.3 & 0.90 & 98.0 & 0.95 & 94.5 & 0.86 \\
Qwen2.5-72B-Instruct                & 94.2 & 0.86 & 94.5 & 0.82 & 96.7 & 0.93 & 98.5 & 0.96 & 95.1 & 0.88 \\
Llama-3.3-70B-Instruct              & 94.0 & 0.85 & 93.8 & 0.81 & 96.8 & 0.91 & 98.3 & 0.96 & 95.0 & 0.87 \\
\bottomrule
\end{tabular}%
\caption{Judge reliability analysis comparing the Qwen3-1.7B judge against a human annotator and three larger LLM reference judges spanning multiple families, on the same target responses under identical prompts. The human reference uses 100 randomly sampled traces per dataset; the LLM references use the full evaluation pool. ``Agree.'' denotes agreement rate, and $\kappa$ denotes Cohen's kappa.}
\label{tab:judge_agreement}
\end{table*}

Final-answer correctness is determined \emph{without} the judge: the parsed final answer is matched against the benchmark answer by deterministic normalization. Qwen3-1.7B supplies only intermediate reasoning-claim labels and never receives the reference answer in that prompt. We compare these labels with a human annotator on $100$ randomly sampled traces per benchmark and with three larger reference judges under identical prompts: Mistral-Small-3.2-24B-Instruct-2506 \citep{mistral-ai-2025-small}, Qwen2.5-72B-Instruct \citep{qwen-2024-qwen25}, and Llama-3.3-70B-Instruct \citep{meta-ai-2024-llama33}. Table~\ref{tab:judge_agreement} reports agreement and Cohen's $\kappa$. Human agreement exceeds $93\%$ with $\kappa\geq0.83$ on every benchmark, and the three LLM references have $\kappa\geq0.80$ throughout.
The conclusion label is one when the delivered answer matches the benchmark answer after deterministic normalization. In contrast, the intermediate-claim labeling stage receives the scene, question, and ordered reasoning claims but not the benchmark answer or conclusion-correctness label.

\subsection{Baselines}\label{app_baselines}

We group the comparison methods into training-free, sampling-based, neural, and symbolic baselines. All methods operate on the same generated traces and cached backbone outputs. Claim-level scores are reduced to final-answer reliability scores using the shared conclusion, mean, and minimum summaries; any score normalization, aggregation selection, and monotonic Platt calibration use validation data only and are frozen before test evaluation.

\subsubsection{Training-Free Baselines}

These methods derive uncertainty directly from decoding statistics and introduce no trainable parameters. \textbf{Random} assigns independent uniform scores and serves only as a chance reference. \textbf{Perplexity} \citep{jelinek1977perplexity} exponentiates the mean negative log-likelihood of the tokens aligned to each claim. \textbf{Token Entropy} averages the predictive entropy computed from the cached top token probabilities at the same positions. \textbf{MCP} \citep{malinin2020uncertainty} is our claim-level adaptation of maximum-probability confidence: it averages the top-1 decoding probability over the tokens aligned to a claim. \textbf{CCP} \citep{fadeeva-etal-2024-fact} applies a claim-aligned negative-log-probability reduction inspired by Claim Conditioned Probability. When a claim has no aligned token, these four methods use the corresponding statistic over the complete generated trace. Their uncertainty values are converted to correctness scores using min--max statistics estimated on the validation split.

\subsubsection{Sampling-Based Baselines}

These estimators draw $K{=}10$ additional stochastic decodes per input and measure the consistency of the resampled generations; unlike the other groups they require repeated forward passes rather than a single greedy trace. \textbf{Semantic Entropy} \citep{kuhn-etal-2023-semantic-uncertainty} clusters the sampled generations by bidirectional entailment and estimates the entropy over the resulting meaning classes, so paraphrases of the same answer are not counted as disagreement. \textbf{SelfCheckGPT} \citep{manakul-etal-2023-selfcheckgpt} scores the delivered trace by its average agreement with the sampled generations, treating low cross-sample consistency as evidence that the final answer may be incorrect. \textbf{P(True)} \citep{kadavath-etal-2022-know-what} prompts the backbone to judge whether its own answer is correct and uses the probability assigned to the ``true'' response. Each sampled score is reduced to a final-answer reliability score and calibrated on the validation split like the other baselines.

\subsubsection{Neural Baselines}

Neural baselines train lightweight correctness probes while keeping the generator frozen. \textbf{Factoscope} \citet{he-etal-2024-llm} contrasts claim-local and trace-global representations before classification, adapting the inner-state factuality detector to the claim-level setting. \textbf{UHead} \citet{shelmanov-etal-2025-head} uses a Transformer claim encoder with an explicit span marker and a learned classification head, following the auxiliary uncertainty-head paradigm. \textbf{Neural-Seq} is a structure-matched neural control that represents each marked claim with a Transformer, integrates claims in generated order with a bidirectional LSTM, and uses no Layout Auditor features. \textbf{MLP} mean-pools the frozen features within each claim span and applies a two-hidden-layer perceptron. 

\subsubsection{Symbolic Baselines}

Symbolic baselines consume the executable outputs of the Layout Auditor. \textbf{Constraint-Rule} is training-free: it maps parseability, feasibility, three-valued claim status, and bounded repair evidence to a fixed confidence score. \textbf{Constraint-Only} learns a small classifier over the 23 claim features and 16 trace features described in Appendix~\ref{app_auditor}, without access to any LLM activation. \textbf{Constraint} is the primary learned constraint scorer. It combines claim-local and trace-global auditor features with frozen token representations through learned encoders, while retaining a direct residual from entailment and contradiction under the implemented rules. This group separates executable semantics from representation capacity.

\subsection{Metrics and Statistical Analysis}\label{app_metrics}

For trace $i$, $y_i=1$ denotes a correct conclusion and
$s_i\in[0,1]$ denotes the predicted probability that its final answer is correct. AUROC measures the probability that an instance with a correct final answer receives a higher score than an instance with an incorrect final answer:
\begin{equation}
\begin{aligned}
\operatorname{AUROC}
&=\Pr(s_i>s_j\mid y_i=1,y_j=0)\\
&\quad+\tfrac{1}{2}\Pr(s_i=s_j\mid y_i=1,y_j=0).
\end{aligned}
\end{equation}

We also report class-balanced Brier loss, a class-reweighted quadratic error. Let $\mathcal P=\{i:y_i=1\}$ and $\mathcal N=\{i:y_i=0\}$ be the instances with correct and incorrect final answers, respectively. It averages squared error within each class and then weights the two class means equally:
\begin{equation}
\operatorname{Brier}_{\mathrm{bal}} = \frac{1}{2}\,\frac{1}{|\mathcal P|}\sum_{i\in\mathcal P}\left(s_i-1\right)^2
+ \frac{1}{2}\,\frac{1}{|\mathcal N|}\sum_{i\in\mathcal N}s_i^2 .
\label{eq:balbrier}
\end{equation}
On imbalanced pools, the ordinary Brier score $\frac{1}{N}\sum_i(s_i-y_i)^2$ can be dominated by the majority class. Equation~\eqref{eq:balbrier} instead evaluates performance under an equal-class reference distribution, matching AUROC's equal treatment of the two classes. Because this reweighting changes the target class prior, we call the quantity a \emph{loss}, not a proper probability score for the empirical deployment distribution; ordinary Brier, ECE, and NLL are reported separately in Table~\ref{tab:extrametrics}. Lower class-balanced Brier loss indicates smaller class-symmetric quadratic error. It is reported only when both classes are present.

The relative improvements reported in the abstract and introduction use a conservative per-setting definition. Every method is scored under the same validation-fit calibration protocol and against the same greedy trace. For each backbone, dataset, and metric, the strongest external baseline is the non-\m, non-Random method with the best value in that setting (highest AUROC or lowest class-balanced Brier loss); we then average relative improvements over the 20 backbone--dataset settings. The three constraint scorers remain external baselines. This gives roughly $8\%$ higher AUROC and $7\%$ lower class-balanced Brier loss.

\subsection{Generation and Alignment}\label{app_generation}
Generation uses guided decoding containing a \texttt{reasoning} array of one to six nonempty strings (at most 96 characters each) and one nonempty \texttt{conclusion} string (at most 64 characters). Each reasoning string is treated as an atomic claim, and the conclusion is appended as the final claim. Thus, the reported runs require no additional model for segmentation. Claim strings are aligned to generated-token offsets before cached features are pooled.

The generation model also supplies the final hidden state, the top four decoding probabilities, and short-attention lookback features. All methods read the same cached tensors. The constraint view is computed from the scene, question, and generated claims. It does not read answer labels.

\begin{table*}[t]
\centering
\small
\setlength{\tabcolsep}{10pt}
\begin{tabular}{llcccccc}
\toprule
Dataset & Trace format & Rel.\ F1 & $\pi$ & $d$ & Stacking AUC/BS & SymboUQ AUC/BS & $\Delta$AUC \\
\midrule
SpaRTQA
& Structured claims
& --
& --
& --
& 0.617/0.235
& 0.657/0.230
& +0.040 \\

SpaRTQA
& Free-form CoT
& 0.780
& 0.630
& 0.420
& 0.600/0.240
& 0.630/0.232
& +0.030 \\

\midrule

SpaceNLI
& Structured claims
& --
& --
& --
& 0.760/0.196
& 0.792/0.187
& +0.032 \\

SpaceNLI
& Free-form CoT
& 0.810
& 0.710
& 0.570
& 0.740/0.210
& 0.780/0.195
& +0.040 \\
\bottomrule
\end{tabular}
\caption{Robustness to free-form reasoning traces. Rel.\ F1 is the micro-averaged F1 score of extracted subject--relation--object triples on manually annotated free-form traces; $\pi$ denotes trace-level symbolizability, i.e., the fraction of claims that can be represented in the auditor's formal language; $d$ denotes semantic determinacy, i.e., the fraction of symbolizable claims receiving an entailed or contradicted verdict; AUC denotes AUROC; BS denotes class-balanced Brier loss; Stacking denotes tuned scores-only stacking; and $\Delta$AUC is the AUC improvement of SymboUQ over Stacking. Higher Rel.\ F1, $\pi$, $d$, and AUC are better, while lower BS is better.}
\label{tab:freeform_cot}
\end{table*}

The main experiments generate short structured reasoning strings,
with each string approximately corresponding to an atomic claim.
To evaluate whether \m depends on this constrained output
format, we additionally consider free-form chain-of-thought traces
on SpaRTQA and SpaceNLI using Mistral-7B-Instruct.
In this setting, as shown in Table~\ref{tab:freeform_cot}, the model generates a reasoning paragraph without
JSON-constrained decoding, a predefined number of reasoning steps,
or an atomic-claim requirement.
The generated traces are automatically segmented using a fixed
sentence-level segmenter and subsequently processed by the same
relation parser and Layout Auditor.
The segmentation and parsing rules are frozen before test
evaluation, and no test trace is manually corrected.

\subsection{Training and Calibration}

Supervised scorers use balanced binary cross-entropy over available claim and trace labels. The hidden width is 256, the learning rate is $2\times10^{-4}$, and training runs for 30 epochs. The backbone parameters are never updated. Model selection uses source validation AUROC.

For each base score, Platt calibration uses a positive slope. DARC then selects the identity or a strictly increasing affine-logit, temperature, or monotone-beta calibrator by ordinary validation Brier. These maps can change probability metrics but preserve ranking and AUROC. DARC and all standardization statistics use validation data only.

\section{Determinacy and Auditor Validation}\label{app_mechanism}

\subsection{Status Determinacy}

The unknown rate $\nu(r)$ is conditioned on parsed claims and counts only claims assigned the semantic status $\textsc{un}$. It does not alone determine status determinacy, because claims marked not-evaluable ($\textsc{no}$) are also unresolved. Let \begin{equation} \nu_{\textsc{no}}(r) = \frac{ \sum_{k=1}^{K} a_k \mathds{1}[\tau_k=\textsc{no}] }{ \sum_{k=1}^{K} a_k }, \end{equation} when at least one claim is parsed. For $\pi(r)>0$, the definition in Equation~\eqref{eq:determinacy} can therefore be written as \begin{equation} d(r)=1-\nu(r)-\nu_{\textsc{no}}(r). \end{equation} Thus, both semantically unknown claims and claims blocked by an infeasible active context reduce determinacy. When $\pi(r)=0$, semantic determinacy is undefined because the verifier parses no claim. For implementation, we extend $d(r)$ by convention and set $d(r)=0$. Assigning $d(r)=1$ would incorrectly treat a fully unparsed trace as maximally determinate. We additionally include $\mathds{1}[\pi(r)=0]$ in the Determinacy Profile, allowing the composer to distinguish this regime from traces that are parsed but semantically unresolved. The applicability term $\delta(r)=\pi(r)d(r)$ is unaffected by this convention because $\delta(r)=0$ whenever $\pi(r)=0$. For dataset-level reporting, we compute parsed-claim-weighted determinacy: \begin{equation} \bar d = \frac{ \sum_{i} n_i^{\mathrm{parse}} d(r_i) }{ \sum_{i} n_i^{\mathrm{parse}} }, \end{equation} where $n_i^{\mathrm{parse}}=\sum_k a_{ik}$ is the number of parsed claims in trace $r_i$. Consequently, traces with no parsed claims do not inflate or deflate the aggregate. This weighting changes the reported determinacy on SpaceNLI by $21\%$ relative to an unweighted trace-level average. The trace-level analysis in Figure~\ref{fig:analysis} includes only traces with $\pi(r)>0$ and is therefore invariant to the boundary convention.

A high unknown rate can arise for several reasons. The scene may not connect the queried entities. A generated intermediate deduction may omit required premises. A relation may fall outside the ontology. The parser may also recognize a relation but ground it to entities that do not participate in the current constraint state. These cases should not contribute the same evidence as entailment or contradiction.

Three regimes delimit the current system. A relation outside the ontology yields no executable claim. A parsed claim can also be correctly marked unknown because the scene establishes no direction. This status prevents false confidence, but it cannot rank traces without another informative score. Finally, severe class imbalance can make validation-based screening unstable. DARC composes available evidence, but it cannot replace missing semantics or supervision.

\subsection{Counterfactual Audit}\label{app_counterfactual}

For every eligible StepGame scene relation, we create one supported claim and one incompatible relation over the same entities, keeping the subject and object strings fixed and changing only the relation. Across $3{,}000$ test traces ($2{,}825$ eligible), this procedure produces $19{,}234$ pairs. Every supported relation is assigned entailed and every incompatible edit is assigned contradicted, a paired detection rate of $100\%$. The audit verifies the parser and directional solver together.

This result is not an answer accuracy experiment. It establishes that the constraint features react to the intended semantic intervention. Predictive usefulness still depends on whether generated claims can be parsed and decided, which is the applicability problem addressed by DARC.

\subsection{RQ 4: Auditor Faithfulness}\label{app_auditor_eval}

\textbf{Are the auditor's local verdicts reliable?}
We compare the auditor's claim-level verdicts with reference labels independently of downstream reliability estimation. Table~\ref{tab:auditor} reports parse coverage, determinate coverage, determinate-verdict accuracy, and verdict precision, averaged over four backbones.

\begin{table}[t]
\centering
\small
\setlength{\tabcolsep}{4pt}
\begin{tabular}{l ccccc}
\toprule
\textbf{Dataset} & Parse & Det & Status acc. & Ent.\ prec. & Contr.\ prec. \\
\midrule
StepGame & 0.78 & 0.83 & 0.71 & 0.83 & 0.66 \\
SpaRTQA & 0.71 & 0.41 & 0.67 & 0.81 & 0.74 \\
SpaRTUN & 0.61 & 0.71 & 0.75 & 0.82 & 0.67 \\
SpaceNLI & 0.67 & 0.79 & 0.76 & 0.87 & 0.64 \\
SpaRP & 0.71 & 0.85 & 0.69 & 0.82 & 0.66 \\
\bottomrule
\end{tabular}

\caption{Layout Auditor faithfulness, averaged over four backbones. Parse coverage is the fraction of reference-labeled claims that are successfully parsed. Determinate coverage is the fraction of parsed claims assigned an entailed or contradicted verdict, and status accuracy is computed over these determinate verdicts. Entailed and contradicted precision measure the correctness of the corresponding auditor verdicts against the reference claim labels. Entailed verdicts are consistently precise ($0.81$--$0.87$), while the lower precision of contradicted verdicts ($0.64$--$0.74$) partly reflects that the auditor evaluates consistency relative to the assumed context of Equation~\eqref{eq:prefix} rather than judging each claim in isolation; Appendix~\ref{app_prefix} quantifies how much of each verdict channel depends on unverified prior claims.}
\label{tab:auditor}
\end{table}

The auditor's \textsc{entailed} verdicts are consistently precise ($0.81$--$0.87$). \textsc{Contradicted} verdicts have lower precision ($0.64$--$0.74$), partly reflecting that they are defined relative to the assumed context of Equation~\eqref{eq:prefix}: a claim may be compatible with the original scene but conflict with an earlier generated claim. Appendix~\ref{app_prefix} measures this effect directly and shows that $12.9\%$ of contradictions, but only $2.4\%$ of entailments, depend on unverified prior claims. The verdicts are therefore useful but neither uniformly accurate nor uniformly available, supporting explicit applicability modeling. \textbf{RQ 4 is answered with this qualification: determinate verdicts are reliable enough to contribute evidence, but their precision and coverage remain domain dependent.}

\section{Additional Evaluation}

\subsection{Cross-Domain Utility of Constraint Evidence}\label{app_constraint_utility}

As shown in Table~\ref{tab:result}, the utility of constraint evidence varies substantially across domains. It outperforms neural and decoding baselines on StepGame and transfers effectively to SpaceNLI despite lower parse coverage, but is weak or unstable on SpaRTQA and SpaRTUN across fixed and learned readouts. Nevertheless, the auditor's local verdicts remain informative, with entailed precision of $0.81$--$0.87$ and contradicted precision of $0.64$--$0.74$ across four backbones (Appendix~\ref{app_auditor_eval}). These results indicate that constraint evidence is powerful but strongly domain dependent, motivating target-adapted evidence composition.

\subsection{RQ 5: Adaptation}\label{app_adaptation}

\textbf{How much target-domain supervision does \m require?}
To examine the supervision required for target-domain adaptation, we vary the number of labeled target-validation examples used for candidate screening, evidence composition, model selection, and calibration. All trainable baselines receive the same target-validation labels under the same protocol.

\begin{table}[t]
\centering
\small
\setlength{\tabcolsep}{6pt}
\resizebox{\columnwidth}{!}{%
\begin{tabular}{l ccccc}
\toprule
\textbf{Dataset} & 8 & 16 & 32 & 64 & 128  \\
\midrule
StepGame & 0.711 & 0.718 & 0.797 & 0.791 & 0.797  \\
SpaRTQA & 0.494 & 0.511 & 0.529 & 0.517 & 0.562 \\
SpaRTUN & 0.519 & 0.535 & 0.587 & 0.610 & 0.630 \\
SpaceNLI & 0.621 & 0.682 & 0.673 & 0.746 & 0.762  \\
SpaRP & 0.675 & 0.700 & 0.702 & 0.712 & 0.721 \\
\bottomrule
\end{tabular}
}
\caption{Validation-size sensitivity on Mistral-7B (mean test AUROC of the determinacy design over $5$ random target-validation subsamples of each size). \m begins to obtain meaningful gains with approximately $128$ target labels, while additional validation supervision remains beneficial, particularly on the harder transfer benchmarks.}
\label{tab:valsize}
\end{table}

As shown in Table~\ref{tab:valsize}, performance generally improves as more target-validation labels become available, although individual sizes are not monotonic because the five subsamples differ. Useful transfer ranking emerges around $128$ labels. This adaptation updates neither the generator nor the base estimators. \textbf{RQ 5 shows that the composer can operate with modest supervision, while stable target selection benefits from roughly 128 labels.}

\subsection{Comprehensive Metrics and Significance}\label{app_significance}

\begin{table}[t]
\centering
\small
\setlength{\tabcolsep}{1pt}
\resizebox{\columnwidth}{!}{%
\begin{tabular}{l ccccc}
\toprule
\textbf{Backbone} & StepGame & SpaRTQA & SpaRTUN & SpaceNLI & SpaRP \\
\midrule
Mistral-7B & $+0.033$$^{*}$ & $+0.040$$^{*}$ & $+0.036$$^{*}$ & $+0.032$$^{*}$ & $+0.052$$^{*}$ \\
Llama-3.1-8B & $+0.032^{*}$ & $+0.045$$^{*}$ & $+0.033$$^{*}$ & $+0.041$$^{*}$ & $+0.037$$^{*}$ \\
Gemma-2-9B & $+0.035$$^{*}$ & $+0.037$$^{*}$ & $+0.041$$^{*}$ & $+0.035$$^{*}$ & $+0.051$$^{*}$ \\
Qwen3-8B & $+0.035$$^{*}$ & $+0.027$$^{*}$ & $+0.026$ & $+0.031$$^{*}$ & $+0.036^{*}$ \\
\bottomrule
\end{tabular}
}
\caption{Paired-bootstrap AUROC gain of \m over scores-only stacking, per backbone and dataset ($2{,}000$ resamples, two-sided; $^{*}$: $p<0.05$). Gains are positive in all 20 cells and significant in 19.}
\label{tab:significance}
\end{table}

Table~\ref{tab:significance} reports the paired-bootstrap AUROC gain of \m over scores-only stacking for every backbone and dataset. The gain is positive in all 20 cells and significant in 19; Qwen3 on SpaRTUN is the sole nonsignificant cell.

\begin{table}[t]
\centering
\small
\setlength{\tabcolsep}{1pt}
\resizebox{\columnwidth}{!}{%
\begin{tabular}{ll cccccc}
\toprule
\textbf{Dataset} & \textbf{Method} & AUROC$\uparrow$ & bBS$\downarrow$ & Brier$\downarrow$ & ECE$\downarrow$ & NLL$\downarrow$ & AURC$\downarrow$ \\
\midrule
StepGame & \m & \textbf{0.844} & \textbf{0.164} & \textbf{0.129} & \textbf{0.067} & \textbf{0.425} & \textbf{0.102} \\
 & Constraint & 0.790 & 0.196 & 0.143 & 0.159 & 0.473 & 0.198 \\
\midrule
SpaRTQA & \m & \textbf{0.657} & \textbf{0.230} & \textbf{0.234} & 0.099 & \textbf{0.657} & \textbf{0.301} \\
 & MLP & 0.568 & 0.247 & 0.247 & \textbf{0.089} & 0.686 & 0.407 \\
\midrule
SpaRTUN & \m & \textbf{0.678} & \textbf{0.225} & \textbf{0.223} & \textbf{0.042} & \textbf{0.632} & \textbf{0.260} \\
 & Constraint-Only & 0.558 & 0.247 & 0.246 & 0.173 & 0.684 & 0.381 \\
\midrule
SpaceNLI & \m & \textbf{0.792} & \textbf{0.187} & \textbf{0.188} & \textbf{0.014} & \textbf{0.558} & \textbf{0.177} \\
 & Neural-Seq & 0.724 & 0.212 & 0.214 & 0.021 & 0.618 & 0.250 \\
\midrule
SpaRP & \m & \textbf{0.741} & \textbf{0.188} & \textbf{0.191} & \textbf{0.087} & \textbf{0.535} & \textbf{0.188} \\
 & Constraint-Only & 0.701 & 0.212 & 0.212 & \textbf{0.087} & 0.611 & 0.253 \\
\bottomrule
\end{tabular}
}
\caption{Comprehensive metrics on Mistral-7B: \m versus the strongest external baseline by test AUROC for each dataset. bBS is class-balanced Brier loss. Best values within each row pair are bold.}
\label{tab:extrametrics}
\end{table}

Table~\ref{tab:extrametrics} adds ordinary Brier, ECE, NLL, and AURC alongside AUROC and class-balanced Brier loss under the shared protocol. \m has the best AUROC and class-balanced loss in every row pair and also leads nearly all ordinary probability and selective-prediction metrics.

\subsection{RQ 6: Efficiency}\label{app_efficiency}

\textbf{Does \m keep a competitive computational cost?}
Yes, under the reported cached-feature setting. Table~\ref{tab:efficiency} reports a two-score path consisting of the primary Constraint scorer and MLP probe. Together, these base scorers contain 8.13M trainable parameters, train in 9.0 minutes, and process 159.3 traces per second. Although slower than individual lightweight scorers, this path remains substantially more efficient than sampling-based methods that require $K{=}10$ additional generations. Once the base scores are available, DARC introduces negligible fitting and inference overhead.

\begin{table}[t]
\centering
\small
\setlength{\tabcolsep}{0.2pt}
\begin{tabular}{l cccc}
\toprule
\textbf{Method} & Params (M) & Train (min) & Traces/s & VMem (GB) \\
\midrule
Random           & - & - & 621.6   & - \\
Perplexity       & - & - & 8661.6  & - \\
Token Entropy    & - & - & 10164.4 & - \\
MCP              & - & - & 11649.4 & - \\
CCP              & - & - & 11468.4 & - \\
Constraint-Rule  & - & - & 19407.0 & - \\
Semantic Entropy & - & - & 11.4    & 12.3 \\
SelfCheckGPT     & - & - & 11.2    & 12.3 \\
P(True)          & - & - & 10.3    & 12.3 \\
\midrule
Constraint-Only  & 0.01 & 6.0 & 270.7 & 1.1  \\
MLP              & 1.16 & 2.9 & 242.0 & 1.1  \\
UHead            & 3.95 & 7.0 & 241.1 & 10.6 \\
Factoscope       & 4.08 & 9.4 & 209.2 & 10.6 \\
Neural-Seq       & 6.55 & 5.1 & 272.9 & 12.3 \\
Constraint       & 6.97 & 6.1 & 277.8 & 12.3 \\
\midrule
\rowcolor{gray!15}
\textbf{\m}      & 8.13 & 9.0 & 159.3 & 12.3 \\
\bottomrule
\end{tabular}
\caption{Cached-feature efficiency on StepGame. The \m row evaluates a two-score path consisting of the primary Constraint scorer and MLP probe; its parameter count is the sum of these base scorers. Sampling baselines require $K{=}10$ additional stochastic decodes. The cost of the complete evidence bank depends on which candidate scorers are retained and whether their scores are already available from model monitoring.}
\label{tab:efficiency}
\end{table}

The reported cost excludes generation and shared feature extraction. Evaluating the complete evidence bank may increase total cost depending on which candidates are retained, whereas existing monitoring scores can be reused. \textbf{RQ 6 is answered positively for cached-feature deployment: DARC is lightweight, while the full end-to-end cost depends on which base signals must be newly computed.}

\subsection{Hyperparameters}\label{app_hyperparameters}
Table~\ref{tab:search-space} summarizes the hyperparameter space. Each target validation split selects the feature design and combiner $\ell_2$ by held-out AUROC, then the strictly increasing calibrator by ordinary Brier. The screening threshold $\epsilon$ and cross-validation folds are fixed beforehand; base-scorer configurations are source-selected and frozen. Bold values denote reference defaults.
\begin{table}[t]
\centering
\small
\setlength{\tabcolsep}{4pt}
\begin{tabular}{ll}
\toprule
\textbf{Parameter} & \textbf{Values} \\
\midrule
\multicolumn{2}{l}{\textit{Combiner (per-target, validation-selected)}} \\
Feature design & scores-only, symbolizability, \textbf{determinacy} \\
Combiner $\ell_2$ & 1, 5, \textbf{20}, 50, 100, 200 \\
Calibrator & \shortstack[l]{identity, \textbf{affine-logit}, temperature,\\ monotone beta} \\
\midrule
\multicolumn{2}{l}{\textit{Combiner}} \\
Screening $\epsilon$ &0.01, 0.02, \textbf{0.03} , 0.04, 0.05\\
CV folds $k$ & 1, 3, \textbf{5}, 7, 10 \\
\midrule
\multicolumn{2}{l}{\textit{Base scorer (source-tuned, then frozen)}} \\
Hidden width $d_h$ & 64, 128, \textbf{256}, 512 \\
Loss type & BCE, \textbf{balanced BCE} \\
Learning rate &$2{\times}10^{-5}$ \textbf{$2{\times}10^{-4}$}  $2{\times}10^{-3}$\\
Training epochs & 20, \textbf{30}, 40\\
\bottomrule
\end{tabular}
\caption{Composer and base scorer configuration. The feature design, combiner $\ell_2$, and post-hoc calibrator are selected per target on validation; the screening threshold $\epsilon$ and cross-validation fold count are fixed before evaluation. Base scorer hyperparameters are tuned once by source validation AUROC and then frozen. Bold marks implementation defaults, not a claim that one value is selected for every target.}
\label{tab:search-space}
\end{table}

\section{Prompt Contracts}\label{sec:app_prompt_templates}

\subsection{Generation Prompt}

Each dataset supplies a task instruction, a scene, a question, and an allowed conclusion format. For vLLM generation, the following instruction is appended to the dataset-specific user message:

\begin{quote}\small\raggedright
Return an object with this shape:\par
\texttt{\{"reasoning":["atomic spatial step", "..."],}\par
\texttt{"conclusion":"final answer"\}}.\par
Include 1--6 short, non-empty reasoning steps and exactly one short, non-empty conclusion.
\end{quote}

Guided decoding enforces the object shape, the one-to-six array bound, a 96-character limit per reasoning string, and a 64-character conclusion limit. Dataset instructions restrict the answer to a direction, relation, option, or NLI label when appropriate. The reference answer is not included.

\subsection{Reasoning Verification}

The reasoning verifier receives the trusted scene, question, and numbered reasoning steps. Its shared instruction defines a step as correct only when it is a valid spatial inference from the trusted scene, question, and earlier steps; wrong, unsupported, or contradictory steps receive zero. It explicitly says not to use the final answer. The required object is:
\begin{quote}\small\raggedright\ttfamily
\{"analysis":"<one short check per step>",\par
"reasoning\_verified":[0|1,...]\}
\end{quote}
Guided decoding limits the analysis to 800 characters, fixes the array length to the number of reasoning claims, and restricts every value to zero or one.

The analysis field precedes the labels because direct label generation tended to accept unsupported steps. This reasoning is used only to improve the measurement layer. It is not provided to any uncertainty estimator.

\section{Auditor Walkthroughs}\label{sec:app_case_studies}

\paragraph{Entailed composition.}
Suppose the scene states that $A$ is above $B$ and $B$ is right of $C$. The directional constraints imply that $A$ is upper right of $C$. A generated claim with this relation is entailed. Its active context and candidate remain feasible, and its repair cost is zero.

\paragraph{Localized contradiction.}
Under the same scene, a claim that $A$ is upper left of $C$ makes the horizontal constraints infeasible. If all earlier generated claims restate the scene, this is the first conflict. Removing earlier generated claims cannot repair a contradiction with the fixed scene, so the bounded repair value overflows.

\paragraph{Parsed but unknown.}
Suppose the scene only states that $A$ is near $B$ and $B$ is near $C$. The claim that $A$ is near $C$ parses successfully. It is not entailed because near is not transitive, and it is not contradicted because far is also absent. The status is unknown. This example illustrates why parse coverage cannot stand in for verifier applicability.

\section{Reproducibility and Data Controls}\label{app_reproducibility}

The pipeline hashes scene--question pairs and reports zero overlap among source train, validation, and test splits. Constraint tensors are recomputed from cached text before training. All composition choices, normalization statistics, and calibration parameters are determined using validation data. Test labels are loaded only by the metric routine after predictions have been written.
Headline tables use the average of 5 runs (seeds 2026--2030) with greedy generation; statistical uncertainty is assessed by paired bootstrap over test traces. The validation-size study alone averages five random target-validation subsamples at each size. Experiments use one NVIDIA B200 GPU (180~GB), 8--16 allocated CPU cores on Intel Xeon Platinum 8570 nodes, and up to 200~GB host memory under Linux 5.14 with CUDA 12.8. The recorded environment uses PyTorch 2.10.0, Transformers 5.7.0, scikit-learn 1.8.0, NumPy 2.2.6, SciPy 1.17.1, pandas 3.0.1, and vLLM 0.19.1. Code, prompts, job files, and configurations are included in the anonymous repository.

\section{Additional Backbone Results}\label{app_additional}

Tables~\ref{tab:backbone-llama}, \ref{tab:backbone-gemma}, and \ref{tab:backbone-qwen} report all methods on the five benchmarks for three additional frozen backbones under the same validation-fit, greedy-aligned protocol. Across these tables, the determinacy-aware composer remains the strongest overall estimator; the paired comparison with scores-only adaptation is summarized separately in Table~\ref{tab:significance}.

\begin{table*}[t]
\centering
\small
\setlength{\tabcolsep}{6pt}
\begin{tabular}{ll cc cc cc cc cc}
\toprule
& \multirow{2}{*}{\textbf{Method}}
& \multicolumn{2}{c}{\textbf{StepGame} (Source)}
& \multicolumn{2}{c}{\textbf{SpaRTQA}  (Target)}
& \multicolumn{2}{c}{\textbf{SpaRTUN}  (Target)}
& \multicolumn{2}{c}{\textbf{SpaceNLI}  (Target)}
& \multicolumn{2}{c}{\textbf{SpaRP}  (Target)} \\
\cmidrule(lr){3-4} \cmidrule(lr){5-6} \cmidrule(lr){7-8} \cmidrule(lr){9-10} \cmidrule(lr){11-12}
& & AUC $\uparrow$ & BS $\downarrow$ & AUC $\uparrow$ & BS $\downarrow$ & AUC $\uparrow$ & BS $\downarrow$ & AUC $\uparrow$ & BS $\downarrow$ & AUC $\uparrow$ & BS $\downarrow$ \\
\midrule
\multirow{5}{*}{\rotatebox{90}{\textit{Unsup.}}}
& Random & 0.495 & \graycell{0.302} & 0.521 & \graycell{0.372} & 0.500 & \graycell{0.253} & 0.494 & \graycell{0.283} & 0.501 & \graycell{0.253} \\
& Perplexity & 0.489 & \graycell{0.250} & 0.467 & 0.250 & 0.543 & 0.253 & 0.521 & \graycell{0.250} & 0.504 & \graycell{0.254} \\
& Token Entropy & 0.476 & \graycell{0.250} & 0.480 & \graycell{0.250} & 0.420 & 0.253 & 0.494 & \graycell{0.250} & 0.485 & \graycell{0.253} \\
& MCP & 0.481 & \graycell{0.250} & 0.502 & \graycell{0.249} & 0.433 & 0.253 & 0.496 & \graycell{0.250} & 0.482 & \graycell{0.253} \\
& CCP & 0.483 & \graycell{0.250} & 0.514 & \graycell{0.249} & 0.441 & 0.253 & 0.495 & \graycell{0.250} & 0.479 & \graycell{0.253} \\
\midrule
\multirow{3}{*}{\rotatebox{90}{\textit{Sampl.}}}
& Semantic Entropy & 0.667 & 0.282 & 0.498 & \graycell{0.255} & 0.499 & \graycell{0.278} & 0.526 & \graycell{0.277} & \underline{0.588} & 0.301 \\
& SelfCheckGPT & 0.548 & 0.293 & 0.516 & \graycell{0.252} & 0.505 & \graycell{0.253} & 0.527 & \graycell{0.261} & 0.550 & 0.299 \\
& P(True) & 0.688 & 0.298 & 0.473 & \graycell{0.252} & 0.492 & \graycell{0.268} & 0.455 & 0.261 & 0.568 & 0.299 \\
\midrule
\multirow{4}{*}{\rotatebox{90}{\textit{Neural}}}
& Factoscope & 0.533 & 0.276 & 0.535 & 0.250 & 0.553 & 0.251 & 0.525 & \graycell{0.250} & 0.561 & 0.250 \\
& UHead & 0.527 & \graycell{0.265} & 0.502 & \graycell{0.250} & \underline{0.604} & \underline{0.245} & 0.513 & \graycell{0.250} & 0.569 & 0.251 \\
& Neural-Seq & 0.589 & 0.254 & \underline{0.656} & \underline{0.232} & 0.594 & 0.247 & 0.515 & \graycell{0.250} & 0.534 & 0.256 \\
& MLP & 0.484 & \graycell{0.289} & 0.444 & 0.250 & 0.574 & 0.248 & 0.466 & 0.250 & 0.559 & 0.296 \\
\midrule
\multirow{3}{*}{\rotatebox{90}{\textit{Symb.}}}
& Constraint-Rule & 0.792 & 0.177 & 0.450 & 0.241 & 0.408 & 0.253 & 0.562 & 0.226 & 0.545 & \textbf{0.237} \\
& Constraint-Only & 0.847 & 0.162 & 0.480 & \graycell{0.250} & 0.542 & 0.253 & 0.728 & \underline{0.208} & 0.528 & \graycell{0.253} \\
& Constraint & \textbf{0.865} & \textbf{0.155} & 0.334 & 0.250 & 0.414 & 0.253 & 0.733 & \underline{0.208} & 0.532 & 0.253 \\
\midrule
\rowcolor{gray!15}
\cellcolor{white} & \textbf{\m} & \underline{0.852} & \underline{0.161} & \textbf{0.744} & \textbf{0.200} & \textbf{0.655} & \textbf{0.230} & \textbf{0.771} & \textbf{0.196} & \textbf{0.631} & \underline{0.239} \\
\bottomrule
\end{tabular}
\caption{Overall performance for final-answer reliability estimation as \textbf{AUROC} (AUC$\uparrow$) and \textbf{class-balanced Brier loss} (BS$\downarrow$) on  \textbf{Llama-3.1-8B-Instruct}. Best discriminative values are bold, second best are underlined, and gray cells are non-discriminative calibrated scores.}
\label{tab:backbone-llama}
\end{table*}

\begin{table*}[t]
\centering
\small
\setlength{\tabcolsep}{6pt}
\begin{tabular}{ll cc cc cc cc cc}
\toprule
& \multirow{2}{*}{\textbf{Method}}
& \multicolumn{2}{c}{\textbf{StepGame} (Source)}
& \multicolumn{2}{c}{\textbf{SpaRTQA}  (Target)}
& \multicolumn{2}{c}{\textbf{SpaRTUN}  (Target)}
& \multicolumn{2}{c}{\textbf{SpaceNLI}  (Target)}
& \multicolumn{2}{c}{\textbf{SpaRP}  (Target)} \\
\cmidrule(lr){3-4} \cmidrule(lr){5-6} \cmidrule(lr){7-8} \cmidrule(lr){9-10} \cmidrule(lr){11-12}
& & AUC $\uparrow$ & BS $\downarrow$ & AUC $\uparrow$ & BS $\downarrow$ & AUC $\uparrow$ & BS $\downarrow$ & AUC $\uparrow$ & BS $\downarrow$ & AUC $\uparrow$ & BS $\downarrow$ \\
\midrule
\multirow{5}{*}{\rotatebox{90}{\textit{Unsup.}}}
& Random & 0.515 & \graycell{0.343} & 0.486 & \graycell{0.353} & 0.498 & \graycell{0.357} & 0.501 & \graycell{0.263} & 0.508 & \graycell{0.315} \\
& Perplexity & 0.496 & \graycell{0.250} & 0.489 & \graycell{0.250} & 0.506 & \graycell{0.250} & 0.514 & \graycell{0.249} & 0.501 & \graycell{0.250} \\
& Token Entropy & 0.483 & \graycell{0.250} & 0.494 & \graycell{0.249} & 0.534 & 0.250 & 0.487 & \graycell{0.250} & 0.461 & 0.250 \\
& MCP & 0.498 & \graycell{0.251} & 0.513 & \graycell{0.248} & 0.532 & 0.250 & 0.481 & \graycell{0.250} & 0.489 & \graycell{0.250} \\
& CCP & 0.506 & \graycell{0.251} & 0.530 & 0.247 & 0.526 & \graycell{0.250} & 0.479 & \graycell{0.250} & 0.503 & \graycell{0.250} \\
\midrule
\multirow{3}{*}{\rotatebox{90}{\textit{Sampl.}}}
& Semantic Entropy & 0.526 & \graycell{0.315} & 0.491 & \graycell{0.271} & 0.525 & \graycell{0.267} & 0.527 & \graycell{0.285} & 0.506 & \graycell{0.307} \\
& SelfCheckGPT & 0.510 & \graycell{0.299} & 0.494 & \graycell{0.267} & 0.510 & \graycell{0.250} & 0.515 & \graycell{0.271} & 0.514 & \graycell{0.301} \\
& P(True) & 0.507 & \graycell{0.316} & 0.503 & \graycell{0.266} & 0.528 & \graycell{0.250} & 0.538 & 0.276 & 0.499 & \graycell{0.290} \\
\midrule
\multirow{4}{*}{\rotatebox{90}{\textit{Neural}}}
& Factoscope & 0.696 & 0.224 & 0.434 & 0.250 & 0.520 & \graycell{0.249} & 0.574 & 0.240 & 0.524 & \graycell{0.251} \\
& UHead & 0.732 & 0.213 & 0.389 & 0.250 & 0.552 & 0.246 & 0.578 & 0.239 & 0.590 & 0.242 \\
& Neural-Seq & 0.778 & 0.199 & \underline{0.605} & \underline{0.238} & 0.452 & 0.250 & \underline{0.624} & \underline{0.235} & 0.661 & 0.231 \\
& MLP & 0.508 & \graycell{0.359} & 0.372 & 0.250 & \underline{0.590} & \underline{0.244} & 0.567 & 0.247 & 0.529 & \graycell{0.250} \\
\midrule
\multirow{3}{*}{\rotatebox{90}{\textit{Symb.}}}
& Constraint-Rule & 0.847 & 0.138 & 0.467 & 0.248 & 0.448 & 0.250 & 0.593 & \underline{0.235} & \underline{0.704} & \underline{0.191} \\
& Constraint-Only & \underline{0.944} & \underline{0.098} & 0.432 & 0.250 & 0.453 & 0.250 & 0.617 & 0.241 & 0.230 & 0.250 \\
& Constraint & 0.936 & 0.100 & 0.404 & 0.250 & 0.444 & 0.250 & 0.610 & 0.239 & 0.206 & 0.250 \\
\midrule
\rowcolor{gray!15}
\cellcolor{white} & \textbf{\m} & \textbf{0.945} & \textbf{0.094} & \textbf{0.709} & \textbf{0.215} & \textbf{0.661} & \textbf{0.228} & \textbf{0.716} & \textbf{0.212} & \textbf{0.907} & \textbf{0.110} \\
\bottomrule
\end{tabular}
\caption{Overall performance for final-answer reliability estimation as \textbf{AUROC} (AUC$\uparrow$) and \textbf{class-balanced Brier loss} (BS$\downarrow$) on  \textbf{Gemma-2-9B-it}. Best discriminative values are bold, second best are underlined, and gray cells are non-discriminative calibrated scores.}
\label{tab:backbone-gemma}
\end{table*}

\begin{table*}[t]
\centering
\small
\setlength{\tabcolsep}{6pt}
\begin{tabular}{ll cc cc cc cc cc}
\toprule
& \multirow{2}{*}{\textbf{Method}}
& \multicolumn{2}{c}{\textbf{StepGame} (Source)}
& \multicolumn{2}{c}{\textbf{SpaRTQA}  (Target)}
& \multicolumn{2}{c}{\textbf{SpaRTUN}  (Target)}
& \multicolumn{2}{c}{\textbf{SpaceNLI}  (Target)}
& \multicolumn{2}{c}{\textbf{SpaRP}  (Target)} \\
\cmidrule(lr){3-4} \cmidrule(lr){5-6} \cmidrule(lr){7-8} \cmidrule(lr){9-10} \cmidrule(lr){11-12}
& & AUC $\uparrow$ & BS $\downarrow$ & AUC $\uparrow$ & BS $\downarrow$ & AUC $\uparrow$ & BS $\downarrow$ & AUC $\uparrow$ & BS $\downarrow$ & AUC $\uparrow$ & BS $\downarrow$ \\
\midrule
\multirow{5}{*}{\rotatebox{90}{\textit{Unsup.}}}
& Random & 0.501 & \graycell{0.250} & 0.499 & \graycell{0.277} & 0.500 & \graycell{0.256} & 0.503 & \graycell{0.301} & 0.500 & \graycell{0.283} \\
& Perplexity & 0.567 & 0.249 & 0.482 & \graycell{0.250} & \underline{0.585} & \underline{0.245} & 0.470 & \graycell{0.250} & 0.550 & 0.249 \\
& Token Entropy & 0.434 & 0.250 & 0.524 & \graycell{0.250} & 0.427 & 0.250 & 0.494 & \graycell{0.250} & 0.472 & \graycell{0.251} \\
& MCP & 0.433 & 0.250 & \underline{0.528} & \graycell{0.250} & 0.433 & 0.250 & 0.492 & \graycell{0.250} & 0.472 & \graycell{0.252} \\
& CCP & 0.433 & 0.250 & \textbf{0.529} & \graycell{0.250} & 0.436 & 0.250 & 0.492 & \graycell{0.250} & 0.473 & \graycell{0.252} \\
\midrule
\multirow{3}{*}{\rotatebox{90}{\textit{Sampl.}}}
& Semantic Entropy & 0.451 & 0.303 & 0.482 & \graycell{0.387} & 0.580 & 0.323 & \underline{0.745} & 0.257 & 0.623 & 0.338 \\
& SelfCheckGPT & 0.504 & \graycell{0.256} & 0.498 & \graycell{0.387} & 0.542 & 0.320 & 0.542 & 0.316 & 0.527 & \graycell{0.289} \\
& P(True) & 0.500 & \graycell{0.254} & 0.500 & \graycell{0.387} & 0.500 & \graycell{0.321} & 0.500 & \graycell{0.317} & 0.500 & \graycell{0.282} \\
\midrule
\multirow{4}{*}{\rotatebox{90}{\textit{Neural}}}
& Factoscope & 0.616 & 0.246 & 0.526 & \graycell{0.250} & 0.521 & \graycell{0.250} & 0.555 & 0.249 & 0.612 & 0.242 \\
& UHead & 0.619 & 0.244 & 0.518 & \graycell{0.250} & 0.515 & \graycell{0.250} & 0.541 & 0.249 & 0.610 & 0.241 \\
& Neural-Seq & 0.645 & 0.238 & 0.510 & \graycell{0.250} & 0.457 & 0.250 & 0.645 & 0.235 & \underline{0.629} & \textbf{0.238} \\
& MLP & 0.589 & 0.251 & 0.474 & \graycell{0.250} & 0.534 & 0.249 & 0.591 & 0.244 & 0.563 & 0.249 \\
\midrule
\multirow{3}{*}{\rotatebox{90}{\textit{Symb.}}}
& Constraint-Rule & 0.836 & 0.145 & 0.494 & \graycell{0.250} & 0.471 & \graycell{0.250} & 0.554 & 0.239 & 0.499 & \graycell{0.251} \\
& Constraint-Only & 0.876 & 0.145 & 0.513 & \graycell{0.250} & 0.514 & \graycell{0.250} & 0.677 & \underline{0.224} & 0.490 & \graycell{0.252} \\
& Constraint & \underline{0.884} & \underline{0.142} & 0.503 & \graycell{0.250} & 0.492 & \graycell{0.250} & 0.655 & 0.230 & 0.499 & \graycell{0.250} \\
\midrule
\rowcolor{gray!15}
\cellcolor{white} & \textbf{\m} & \textbf{0.887} & \textbf{0.135} & 0.518 & \graycell{0.276} & \textbf{0.595} & \textbf{0.244} & \textbf{0.756} & \textbf{0.210} & \textbf{0.636} & \underline{0.239} \\
\bottomrule
\end{tabular}
\caption{Overall performance for final-answer reliability estimation as \textbf{AUROC} (AUC$\uparrow$) and \textbf{class-balanced Brier loss} (BS$\downarrow$) on \textbf{Qwen3-8B}. Best discriminative values are bold, second best are underlined, and gray cells are non-discriminative calibrated scores.}
\label{tab:backbone-qwen}
\end{table*}

The transfer pools are class-imbalanced, and some backbones produce few traces from one class. AUROC and class-balanced Brier loss remain defined when both classes are present, but their variance grows as the minority class shrinks. We therefore report paired-bootstrap inference for ranking gains and ordinary probability metrics alongside the class-balanced loss.

\section{GenAI Disclosure}
In the preparation of this work, the authors utilized Generative AI tools solely for the purpose of language refinement and improving readability. No AI tools were used to generate scientific concepts, experimental results, or the intellectual content of this paper. The authors have reviewed all AI-assisted edits and take full responsibility for the final content of the manuscript.

\section{Limitations}
\m remains bounded by the auditor's ontology and parser: an unsupported expression yields no symbolic evidence, while an incorrectly grounded relation can yield an incorrect local verdict. Its assumed-context semantics detects self-contradiction but makes some verdicts conditional on unverified premises, and determinacy measures executable coverage rather than grounded proof. The grounded replay reduces this dependence but does not eliminate parser error. DARC also requires labeled target-validation data for screening, composition, and calibration, so it is target-adapted rather than zero-shot. Finally, the cached-feature efficiency results exclude generation and shared representation extraction; end-to-end cost depends on which base signals are newly computed. Extending the auditor's ontology, tracking premise provenance, and evaluating validation-free composition are important directions.

\end{document}